\documentclass[10pt,twocolumn,letterpaper]{article}

\usepackage{iccv}
\usepackage{times}
\usepackage{epsfig}
\usepackage{graphicx}
\usepackage{amsmath}
\usepackage{amssymb}

\usepackage{bm}
\usepackage{xcolor}
\usepackage{float}
\usepackage{mathtools}
\usepackage{caption}
\usepackage{subcaption}
\usepackage{multirow}
\usepackage{makecell}
\usepackage{lipsum}
\usepackage{colortbl}
\usepackage{colortab}
\usepackage{placeins}
\usepackage{tikz}
\usepackage{microtype}

\usepackage[export]{adjustbox}
\usepackage{stfloats}  

\usepackage[accsupp]{axessibility}  

\usepackage[breaklinks=true,bookmarks=false]{hyperref}

\iccvfinalcopy 

\def\iccvPaperID{8949} 

\newcommand{\bs}[1]{\boldsymbol{#1}}
\newcommand{\uf}[1]{\underline{#1}}

\DeclareMathOperator*{\mean}{mean}

\DeclareMathOperator*{\E}{\mathbb{E}}
\newcommand{\roverline}[1]{\displaystyle\mathpalette\doroverline{#1}}
\newcommand{\doroverline}[2]{\overline{#1#2}}

\definecolor{airforceblue}{rgb}{0.36, 0.54, 0.66}

\definecolor{orange}{rgb}{0.60, 0.30, 0.0}

\usepackage[normalem]{ulem}

\begin{document}

\title{S-TREK: Sequential Translation and Rotation \\ Equivariant Keypoints for local feature extraction}

\author{
Emanuele Santellani *, 
Christian Sormann *, 
Mattia Rossi {}\textsuperscript{\textdagger},
Andreas Kuhn {}\textsuperscript{\textdagger},
Friedrich Fraundorfer * \\
* Graz University of Technology, {}\textsuperscript{\textdagger} Stuttgart Laboratory 1, SSS-EU, Sony Europe B.V. \\
{\tt\small * name.surname@icg.tugraz.at, {}\textsuperscript{\textdagger} name.surname@sony.com}
}

\maketitle

\begin{abstract}
In this work we introduce S-TREK,
a novel local feature extractor that combines a deep keypoint detector,
which is both translation and rotation equivariant by design,
with a lightweight deep descriptor extractor.
We train the S-TREK keypoint detector within a framework inspired by reinforcement learning,
where we leverage a sequential procedure to maximize a reward directly
related to keypoint repeatability.
Our descriptor network is trained following a ``detect, then describe'' approach,
where the descriptor loss is evaluated only at those locations 
where keypoints have been selected by the already trained detector.
Extensive experiments on multiple benchmarks confirm the effectiveness of our proposed method,
with S-TREK often outperforming other state-of-the-art methods in terms of repeatability 
and quality of the recovered poses,
especially when dealing with in-plane rotations.
\end{abstract}

\section{Introduction}

Being able to find point correspondences between images has been of paramount importance since the 
early days of computer vision.
In fact, a wide range of applications, such as Structure from Motion (SfM) \cite{schoenberger2016sfm, rome}, 
Visual Localization \cite{visual_localization, outdoorvisuallocalization}, 
SLAM \cite{orbslam, slamsurvey}, object recognition \cite{object_recognition} and
object tracking \cite{object_tracking} rely on image-to-image point correspondences.

After decades where SIFT \cite{sift}, SURF \cite{surf}, ORB \cite{orb} and many other hand-engineered feature extractors have
been ubiquitous, the community has recently experienced a fast shift toward learned methods, 
with several ones based on deep architectures \cite{d2-net, superpoint, r2d2, disk}.
While many of these newly proposed methods show remarkable matching performances,
the commonly used multi-layer convolutional architecture
lacks one of the properties that most hand-engineered keypoint detectors have by design: 
rotation equivariance.
Rather unexpectedly, modern detectors show poor performances when the input image undergoes 
in-plane rotations unless specifically trained to handle this transformation.
In order to avoid this pitfall, we take advantage of the recent developments in the field of
group-equivariant networks \cite{escnn, groupequivariantnn} 
and design the keypoint detector of our novel feature extractor method,
named S-TREK, 
to make use of rotation-equivariant convolutional layers.
This makes our keypoints independent of the image orientation by design, 
regardless of the dataset used for training.

\begin{figure}[t]
     \newcommand{\mysize}{0.32\linewidth}
     \centering
     \begin{subfigure}[b]{\mysize}
         \includegraphics[width=\linewidth]{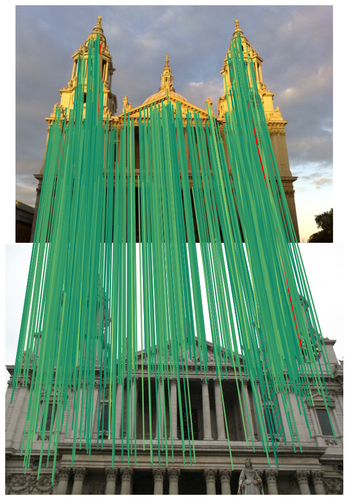}
     \end{subfigure}
     \begin{subfigure}[b]{\mysize}
         \includegraphics[width=\linewidth]{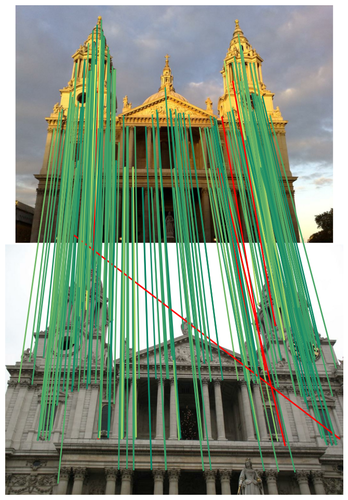}
     \end{subfigure}
     \begin{subfigure}[b]{\mysize}
         \includegraphics[width=\linewidth]{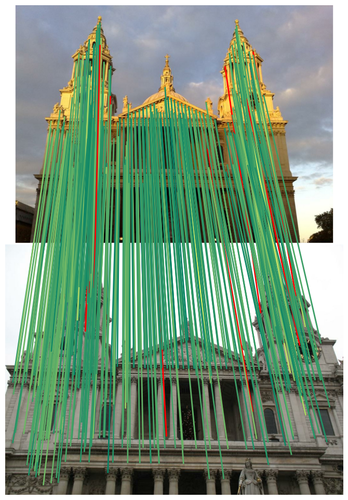}
     \end{subfigure}
     \\
     \begin{subfigure}[b]{\mysize}
         \includegraphics[width=\linewidth]{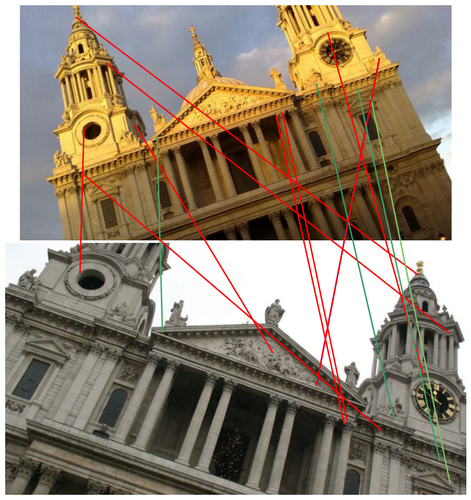}
         \caption{DISK \cite{disk}}
     \end{subfigure}
     \begin{subfigure}[b]{\mysize}
         \includegraphics[width=\linewidth]{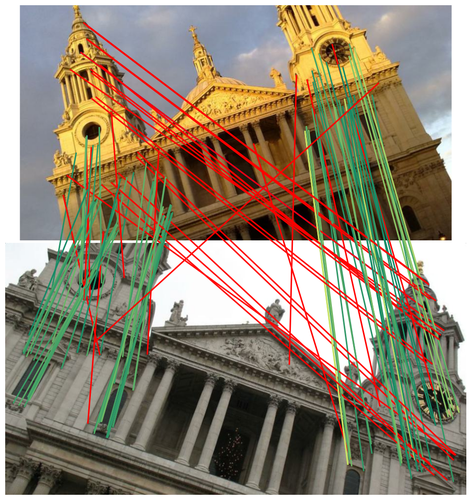}
         \caption{REKD \cite{REKD}}
     \end{subfigure}
     \begin{subfigure}[b]{\mysize}
         \includegraphics[width=\linewidth]{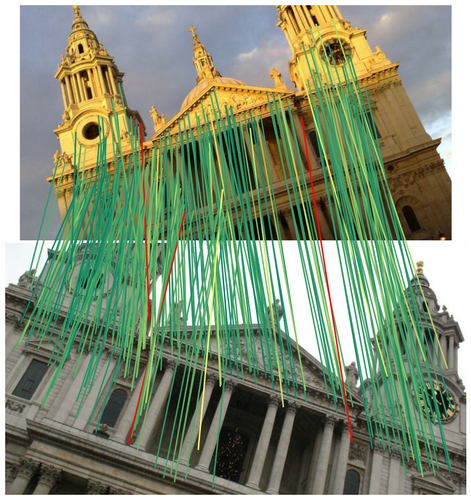}
         \caption{S-TREK (ours)}
     \end{subfigure}
     \caption{
     Qualitative comparison with two state-of-the-art feature extraction methods on the Image Matching Benchmark \cite{imb} (top) and on our $\pm$ 45° rotated version of it (bottom). RANSAC inlier matches are color coded from green to yellow, representing reprojection errors equal to zero and 5px, respectively; the outlier matches are in red.
     }
     \vspace*{-1.0mm}
     \label{fig:imb}
\end{figure}

\begin{figure*}[ht]
\centering
\begin{subfigure}[b]{0.39\linewidth}
    \includegraphics[width=\linewidth]{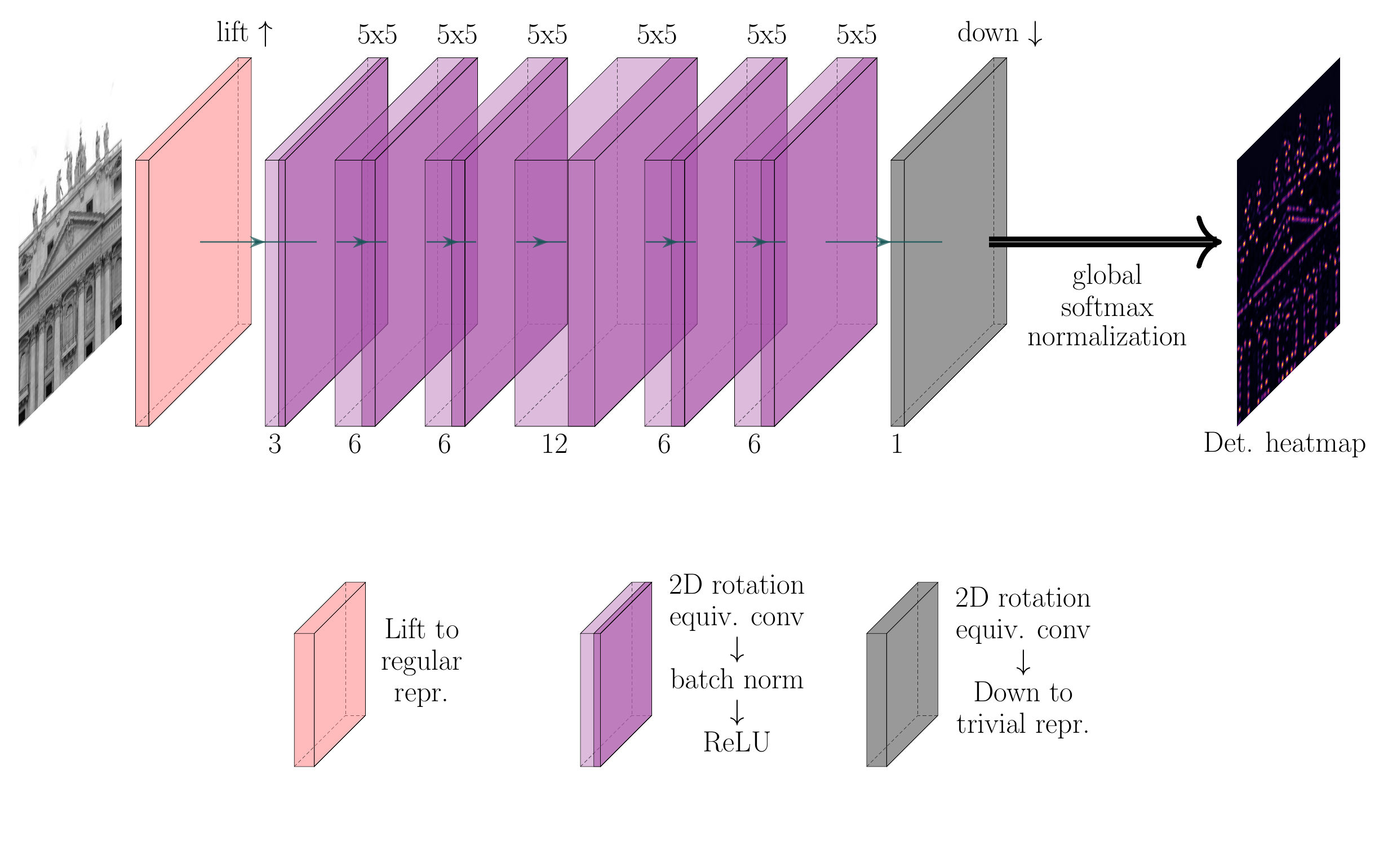}
    \caption{Keypoint detector.}
    \label{fig:architecture-det}
\end{subfigure}
\begin{subfigure}[b]{0.59\linewidth}
    \includegraphics[width=\linewidth]{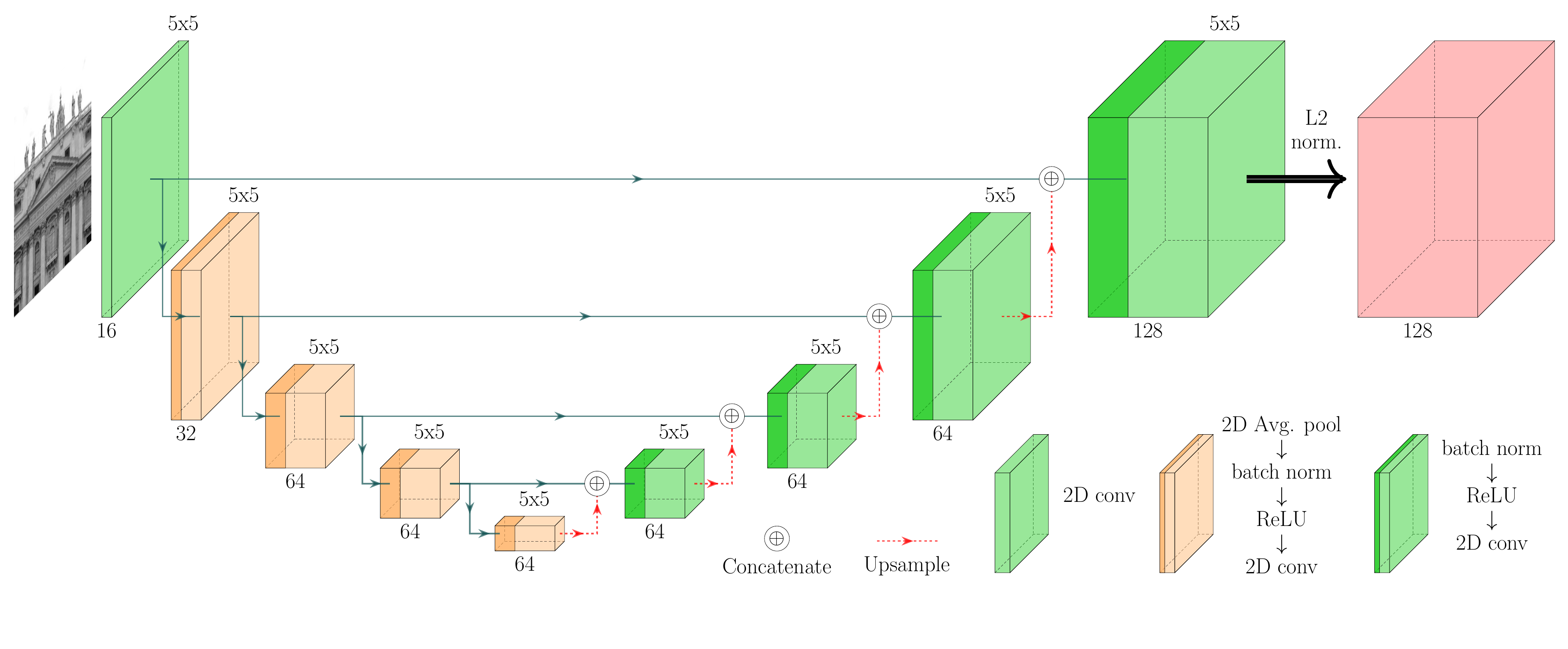}
    \caption{Descriptor extractor.}
    \label{fig:architecture-des}
\end{subfigure}
\caption{Overall architecture of the S-TREK feature extractor.}
\label{fig:architecture}
\vspace*{-1.0mm}
\end{figure*}

The detection of keypoints in an image is a hard selection process,
where a finite set of locations is selected.
Because of the non-differentiability of this process, 
some keypoint detection methods in the literature 
resort to training with with proxy losses, 
which are applied to the entire detection heatmaps at the network output \cite{r2d2, mdnet},
while others train for keypoints and descriptors jointly \cite{d2-net, aslfeat}.
To train directly with the keypoints location, 
we propose a training framework inspired by reinforcement learning,
which permits to maximize a reward formulation directly related to the keypoints repeatability.
Following \cite{disk,reinforced_features_points},
we frame the keypoint detection as a probabilistic process, 
and present a novel \textit{sequential sampling} procedure that does not
suffer from the limitations of their sampling schemes.

In contrast to the more recent \textit{``detect and describe''} approach, 
where keypoint locations and descriptors are learnt jointly
using a shared backbone for both tasks \cite{r2d2, d2-net, disk, mdnet, reinforced_features_points},
we follow a \textit{``detect, then describe''} paradigm
using two different networks for the two tasks,
and train the descriptors only after the detector has been trained.
This allows us to design the two networks to have different properties.

Recent studies have also explored new research directions,
utilizing deep matching architectures like SuperGlue \cite{superglue},
or developing methods able to find pointwise correspondences directly from image pairs such as LoFTR \cite{loftr}.
However, these methods require execution for each individual image pair,
which limits their applicability in scenarios where the computational resources are constrained,
or when dealing with a large number of images.
For these reasons, and the additional benefit of an easier integration in existing systems,
local feature extraction methods remain the most popular approach in many applications.

In summary, our main contributions are:
\begin{itemize}
    \setlength\itemsep{0.0em}
    \item We propose S-TREK, a local feature extractor that combines 
        a deep translation and rotation equivariant keypoint detector with a 
        lightweight descriptor extractor.
        Our network is trained from scratch 
        following a \textit{``detect, then describe''} approach.
    \item We propose a keypoint detector training framework, inspired by reinforcement learning,
        based on a novel reward formulation that maximizes the repeatability metric directly.
        Additionally, we propose a novel sequential keypoint probability sampling strategy
        that overcomes the limitations of previous approaches.
    \item Extensive experiments show that the S-TREK detector achieves state-of-the-art repeatability on multiple benchmarks.
        Moreover, when equipped with our lightweight features extractor network, 
        S-TREK provides features that are well suited for recovering accurate camera poses,
        especially when dealing with in-plane rotation.
\end{itemize}

\section{Related works}

A multitude of local feature extractors have been proposed in the last decades \cite{localfeaturesurvey, localfeaturebenchmark0, localfeaturebenchmark1}.
In the most classical approaches, the local feature extraction task 
is separated into two distinct steps: detecting keypoints and extracting descriptors.
The two tasks are approached by means of specialized algorithms,
obtaining keypoint-descriptor pairs as a result \cite{harris, fast, brief, sift, surf}.
The early deep learning approaches have been designed following the same \textit{``detect, then describe''} approach,
with initial works addressing either keypoint detection \cite{tilde, taskdetector, quadnet, keynet} 
or descriptor extraction from normalized image patches \cite{hardnet, l2-net, sosnet}. 
In \cite{logpolar}, a log-polar sampling scheme is used to extract descriptors which result more robust to scale differences.
Later works propose to design the descriptor network to output a dense volume
where descriptors are sampled at keypoint locations \cite{lift, superpoint, lfnet},
shifting toward a \textit{``detect and describe''} approach 
where the two tasks are trained jointly \cite{d2-net, r2d2, aslfeat, mdnet, disk, reinforced_features_points}.
The rationale behind this shift is keypoint matchability:
a keypoint, regardless of his repeatability, 
could be surrounded by a non-discriminative region (e.g. textureless patch, repetitive structures)
which makes it hard to match its associated descriptor correctly.
While this is true if the descriptors are computed based solely on small patches around 
each keypoint as in the early methods,
increasing the descriptor support regions permits to assign a 
discriminative descriptor to most of the keypoints.
Based on this 
observation, 
we adopt a \textit{``detect, then describe''} approach and
design our method with two distinct networks,
each one with an architecture specialized for the specific task.

This allows us to design our keypoint detector network with rotation equivariant convolutions
as in REKD \cite{REKD}, thus obtaining detections robust to in-plane rotation by design.

To circumvent the non-differentiability of the keypoint selection process, REKD \cite{REKD} uses
a window-based keypoint detection loss inspired by \cite{keynet} that softens the keypoints
selection by means of a spatial softmax, but requires to fix the number of keypoints used during training.
Other recent methods approach the local feature learning 
with training frameworks that draw inspiration from reinforcement learning, 
such as DISK \cite{disk} and Reinforced Feature Points \cite{reinforced_features_points}.
While these methods optimize for both keypoints and descriptors at the same time,
for the aforementioned reasons we focus first on the keypoints, 
postponing the descriptor learning to a later stage.
DISK \cite{disk} models the probabilistic sampling of keypoints by dividing the input image
into a regular grid and sampling one keypoint from each cell. 
However, this approach has some shortcomings. 
For instance, even if a cell contains no stable keypoints, a keypoint will still be sampled from it. 
Additionally, if two valid keypoints are present in the same cell, only one of them can be sampled. 
Moreover, if a keypoint falls on one of the cell edges, it is likely to be sampled by both cells with only a 1px offset.
In Reinforced Feature Points \cite{reinforced_features_points} instead the probabilistic sampling is modeled on top of
an already trained SuperPoint network \cite{superpoint}, sampling multiple times from the whole image
with the danger of repeatedly sampling the same keypoint.
Our approach is to use a \textit{Sequential Sampling} procedure which solves all these issues.
We do not soften the keypoint selection by averaging coordinates like \cite{reinforced_features_points}, but by means of a probabilistic sampling.
We do not require the image to be divided into cells,
as we normalize the whole computed detection heatmap at once and sample from it directly,
thus avoiding all the cell related issues.
We only fix the max number of keypoints sampled during training, but not the minimum one;
our sampling procedure has an early-stopping mechanism where no further keypoints are sampled
if no stable keypoints are left in the image.
Finally, we avoid the double-sampling of the same point 
by applying a \textit{sampling avoidance radius} at each sampled keypoint,
which forces the next sampled keypoint to lay at least N pixels from any already sampled ones.

\section{\fontsize{11}{11}\selectfont Preliminaries on Equivariance and Invariance}
\newcommand{\img}{\mathbb{R}^{H \times W}}

The purpose of this section is to provide the reader with a basic understanding of the \textit{invariance} and \textit{equivariance} properties.
Although we will mostly deal with equivariance in this paper, it is useful to agree on the vocabulary and spend a few words on what invariance means.

Given the sets $X$, $Y$ and a set $G$ of actions ${g: X \rightarrow X}$, a function ${f: X \rightarrow Y}$ 
is invariant with respect to $G$ if, for any ${g \in G, f[g(x)]=f(x)}$.

For example, if $f$ measures the area of a domain in the plane, such a function would be invariant with respect to translations but not with respect to changes of scale.

If the domain and codomain of $f$ are the same, that is $f: X \rightarrow  X$, equivariance with respect to the set of actions $G$ is defined as the property guaranteeing ${f[g(x)]=g[f(x)]}$.

With few approximations,
the frequently used convolutional layer 
can be considered equivariant with respect to translations,
and this property carries over to a network consisting of several layers.
Formally, if ${X = \img}$ is the set of input images, T is the set of translations 
${t: \img \rightarrow \img}$, and $f$ is the operation carried out by our network, the following holds true:
\begin{equation}
    f[t(x)] = t[f(x)], \forall t \in T. 
    \label{eq:equivariance}
\end{equation}
Equivalently, if the input undergoes a translation of an integer number of pixels in the $x$ or $y$ direction, the resulting network output will be translated by the same amount. However, 
the same is not true in general for other geometric transformations that our input might undergo.

In this work, we make use of special convolutional layers that guarantee the equivariance property with respect to the set of in-plane rotations $T_R$ of the input image, 
where the function $f$ represents our keypoint detector deep network $f_\theta$, and $\theta$ represents the learnable parameters of the model.

For the purpose of this work, 
the preliminary introduction we just provided will be sufficient to frame and understand our contribution.
We refer to \cite{escnn, categories_working_mathematician} for further details.

\begin{figure}[t]
     \newcommand{\mysize}{0.49\linewidth}
     \centering
     \begin{subfigure}[b]{\mysize}
         \includegraphics[height=\linewidth]{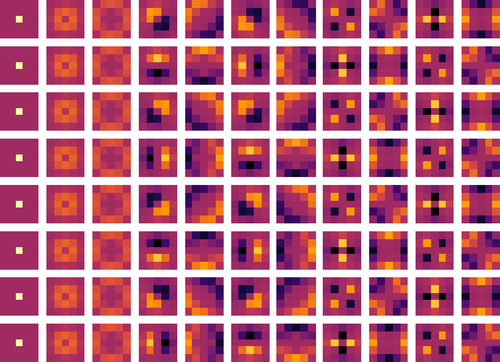}
     \end{subfigure}
     \begin{subfigure}[b]{\mysize}
        \begin{flushright}
         \includegraphics[height=\linewidth]{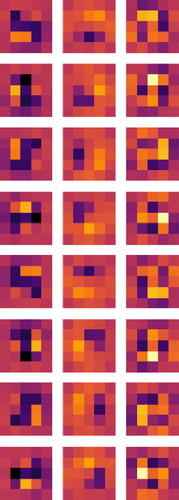}
        \end{flushright}
     \end{subfigure}
     \caption{
      Basis (left) and resulting filters (right) of the first layer of our rotation equivariant keypoint detector. 
      Each column of filters is a linear combination, with learnable weights, 
      of the columns from the basis, and contributes to a single output channel.
     }
     \label{fig:radial_basis}
     \vspace*{-1mm}
\end{figure}

\section{Method}
In contrast to many recent works that train jointly for the two tasks of keypoint detection and descriptor extraction, 
our approach follows the \textit{``detect, then describe''} paradigm.
Our architecture is inherently partitioned into two parts: the \textit{detector} and \textit{descriptor extractor},
which are trained separately.
By adopting this approach, we are able to address the two tasks individually and employ specialized architectures for each of them.

We define a keypoint just as \textit{repeatable} point, 
in other words a point that can be reliably detected again in another image.
This definition also includes points surrounded by textureless areas or that lay on repetitive structures, which are notoriously difficult to match.
To address this challenge, we equip our descriptor network with a very large receptive field 
and train the descriptors only at the exact locations where our keypoints are detected to
avoid wasting descriptor space for non-repeatable regions.
Conversely, we argue that the detector should concentrate solely on local structures, 
designing its backbone with a small receptive field.

Our definition of keypoint requires the detections to remain consistent under any transformations,
whether photometric or geometric, applied to the image.
While CNNs have proven to be easily trained to be robust with respect to photometric distortions,
and translation equivariance is inherent in the convolutional architecture,
robustness against other transformations must be encoded 
in the network's learned weights.
Recent developments in the field of Steerable Filters and Group-equivariant networks \cite{escnn, groupequivariantnn}
have made it possible to effectively embed in-plane rotation equivariance in neural networks.
In a rotation-equivariant convolution (ReCONV), instead of learning the kernel weights like in a standard convolution,
the layer learns a weight for each precomputed basis and generates the convolution kernel
as a weighted basis sum.
This greatly reduces the number of learnable parameters compared to a
standard convolution with the same kernel size.
In Figure~\ref{fig:radial_basis} we show the basis and
and resulting kernels for our first \textit{detector} layer.

\subsection{Detector}
Our \textit{detector} architecture is composed by a series of ReCONV layers.
The first layer transforms the input features into the chosen \textit{regular representation} 
which depends on the encoded rotation cyclicity (i.e. the number of angles at which the basis are computed).
The middle section applies convolutions on the \textit{regular representation}.
Finally, the last layer converts the feature representation back to a single-channel heatmap,
which is then normalized with a \textit{temperature softmax} operation.
This network runs on the input gray-scale image ${\bs{I} \in \img}$ and 
outputs the heatmap ${\bs{D} \in \img}$.
A scheme of the network architecture
is visible in Figure~\ref{fig:architecture-det}.

In order to train the detector,
we propose a simple formulation that applies to the 
extracted keypoints, rather than to the output heatmap $\bs{D}$, 
and aims at maximizing the keypoints repeatability directly.
Typically the keypoint extraction process relies on the $argmax$ operation, 
which prevents back-propagation to the network weights.
In order to overcome this, we draw inspiration from reinforcement learning 
and frame the learning process as an expected reward maximization. 
We adopt a probabilistic approach to sample keypoints and use 
the \textit{policy gradient} \cite{policygradient} algorithm to update the network weights after each training step.

\subsubsection{Reward Computation}
Our reward formulation is very straightforward, given two input images 
${\bs{I}_0, \bs{I}_1 \in \mathbb{R}^{H \times W}}$ and the invertible relation
${g_{0 \rightarrow 1}\colon \mathbb{R}^2 \rightarrow \mathbb{R}^2}$
that projects the coordinate of a keypoint from the first image into the second image,
the reward for the i-th keypoint ${k_i^0}$ in the first image is computed as:
\begin{equation}
    r_{k_i^0} = 
    \begin{cases*}
    d_{\text{max}} - d & if $d \le d_{\text{max}}$ \\
    r_n     & otherwise
    \end{cases*}
\end{equation}
where $d_{\text{max}}$ is a defined \textit{reward radius},
$r_n$ is the \textit{negative reward} assigned to keypoints that are not considered repeatable and
$d$ is the distance between the projected keypoint $k_i^0$ and
the closest keypoints in the second image.
This reward function relates directly to the repeatability measure, 
defined later in Sec~\ref{sec:experiments},
as it is computed in an equivalent way.

\begin{figure}
     \newcommand{\mysize}{0.24\linewidth}
     \centering
     \begin{subfigure}[b]{\mysize}
         \includegraphics[width=\linewidth]{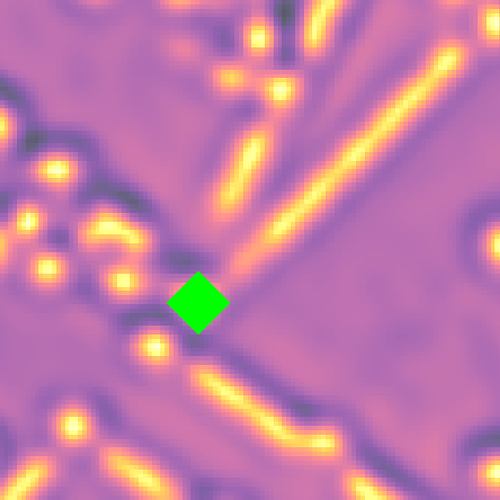}
     \end{subfigure}
     \begin{subfigure}[b]{\mysize}
         \includegraphics[width=\linewidth]{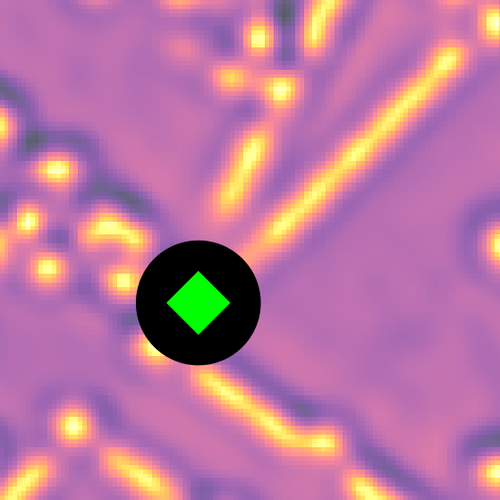}
     \end{subfigure}
     \begin{subfigure}[b]{\mysize}
         \includegraphics[width=\linewidth]{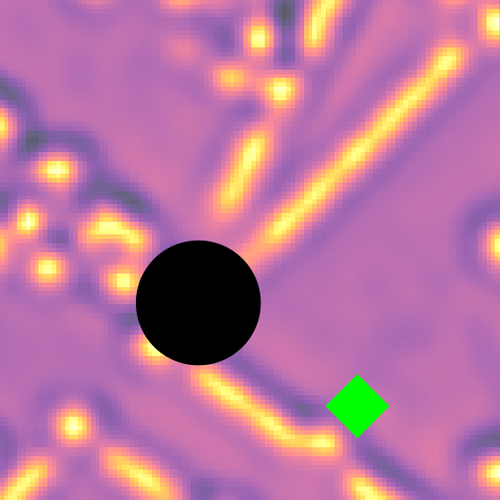}
     \end{subfigure}
     \caption{
     Sequential sampling process. A keypoint is sampled from the 
     weight map obtained by normalizing the heatmap 
     at the keypoint detector output. Then, all the weights in a \textit{sampling avoidance radius} are set to 0. A new keypoint can now be sampled from the updated weight map and the process is iterated. 
     }
     \label{fig:sequential_sampling}
     \vspace*{-1.0mm}
\end{figure}

\subsubsection{Sequential Sampling}
For the \textit{policy gradient} \cite{policygradient} algorithm to work the expected reward must be computed.
In our scenario, this involves a probabilistic selection of the keypoints.

The other methods in the literature \cite{disk, reinforced_features_points}
either divide the image into cells and sample one keypoint for each cell
or sample multiple keypoints independently from the normalized detection map.
The main limitation of the first approach is that
one keypoint is sampled from every cell,
regardless of whether there are two high probability points or none. 
Furthermore, the grid size defines both the spacing between the keypoints
and the total number of sampled keypoints. 
Finally, peaks that lie on the border between two adjacent cells are likely to be sampled in both just with a 1px offset.
The second approach, instead is susceptible to the problem of sampling the same keypoint multiple times.
To tackle both issues, we propose a \textit{sequential sampling} procedure.
First, we normalize the heatmap $\bs{D}$ 
by applying the \textit{softmax} operation with temperature $t$:

\begin{equation}
    \bs{\roverline{D}} = \mathrm{softmax}\left(\frac{\bs{D}}{t}\right).
\end{equation}
Then, we treat $\bs{\roverline{D}}$ as a weight map
and draw N keypoints sequentially with a probability proportional to the corresponding pixel weight.
To avoid sampling both the same keypoint twice or spatially close keypoints, 
after each sampling we set to zero all the weights in a \textit{sampling avoidance radius} around the sampled keypoints.
The sampling process stops when the sum of all the remaining weights is below a certain threshold,
as this suggests that no stable keypoints are left.
This allows the network to tune the number of keypoints sampled during each training iteration automatically.
Figure~\ref{fig:sequential_sampling} shows a graphical representation of this process.

\subsubsection{Weight update}
For each training iteration we run our detector network on a pair of images and 
then sample from the obtained heatmaps the two sets of keypoints $K^0$ and $K^1$.

We compute the reward for each keypoint and estimate the total expected reward as follows:
\begin{equation}
    \E \left[ R(K^0, K^1) \right] = 
    \sum_{k \in K^0} p(k) r_{k} + \sum_{k \in K^1} p(k) r_{k}
\end{equation}
where $p(k)$ refers to the probability of sampling the keypoints $k$,
which is approximated with the value of the normalized heatmap $\bf{\roverline{D}}$ at the coordinate of the keypoint.

Using the \textit{policy gradient} \cite{policygradient} technique we can then estimate the gradient of the expected reward
with respect to the network parameters $\theta$ as follows:
\begin{align}
\begin{split}
    &\nabla_\theta \E \left[ R(K^0, K^1) \right] = \\
    &\mean_{k \in K^0} \left[ \nabla_\theta \log\left(p(k)\right) r_{k} \right] + 
     \mean_{k \in K^1} \left[ \nabla_\theta \log\left(p(k)\right) r_{k} \right]
\end{split}
\end{align}
and use this gradient to optimize the network weights.

\subsection{Descriptor extraction}
We design our \textit{descriptor} network as a four-level U-Net \cite{unet}
with a single convolutional layer per level, 
as shown in Figure~\ref{fig:architecture-des}.
The network is fed with the same grayscale image $\bs{I}$ and outputs
a dense L2-normalized descriptor volume ${\bs{V} \in \mathbb{R}^{H \times W \times d}}$
where $d$ is the descriptor dimension.
The descriptor associated to a detected keypoint is simply read from the volume at the keypoint location.
We employ a hinged triplet loss formulation \cite{hardnet}:
\begin{equation}
   \mathcal{L}_\text{Descriptor} = \mean_{t \in \mathcal{T}} \big[\max(0, m + s_n^t - s_p^t) \big]
   \label{eq:triplet}
\end{equation}
where $t$ is a triplet from the pool $\mathcal{T}$ of all the chosen anchor-positive-negative triplets,
$m$ is the margin, $s_n$ is the anchor-negative score and $s_p$ is the anchor-positive score.
The triplets are built following the \textit{hardest} strategy \cite{hardnet}, 
where we pick the closest non-matching descriptor for each anchor.

\begin{figure}[t]
\begin{center}
\includegraphics[trim={2.5cm 0 2cm 0}, clip, width=\linewidth]{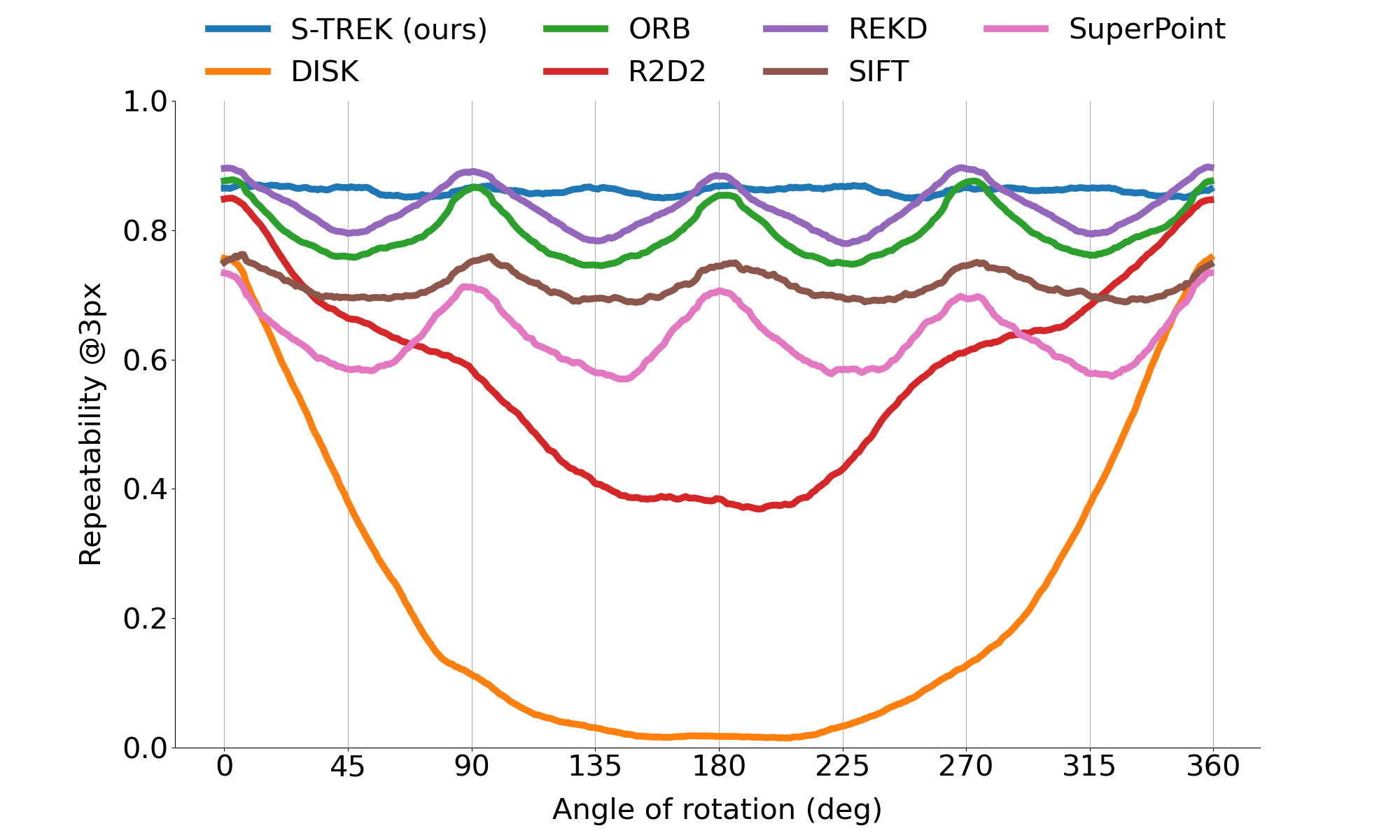}
\end{center}
\vspace{-4.0mm}
\caption{Repeatability as a function of image rotation angle. 
The curves are smoothed with a 15 degrees moving average.}
\label{fig:rotation_experiment}
\vspace{-1.0mm}
\end{figure}

\section{Training Setup}
We train the two networks sequentially on the subset of MegaDepth \cite{MegaDepthLi18} provided by \cite{disk}.
The dataset comprises images, camera poses, camera intrinsics and depth maps of 3D reconstructed touristic locations.
The subset includes a total of 135 scenes and
is the result of removing the scenes with unreliable depths 
and those overlapping with the Image Matching Benchmark \cite{imb}.
We draw random image pairs from the provided list of 10k image pairs per scene.
We rescale each image such that the shortest edge measures 512px 
and crop the other dimension to obtain 512$\times$512 image patches.
For the descriptor training, we apply an additional random rotation 
sampled from a uniform distribution [-30, 30] degrees to each image.

The first training involves only the \textit{detector} network, 
where we use the ESCNN library \cite{escnn} for the ReCONV and set the cyclicity to 8 
(the basis are generated for each $2\pi/8$ rotation).
We train from scratch using the Pytorch framework \cite{pytorch}, 
Adam optimizer \cite{adam} with betas (0.9, 0.999), 
learning rate 1e-4 and batch size 4.
We set the \textit{sampling avoidance radius} to 6px, 
the reward radius $d_{max}$ to 3px,
the number of max sampled points to 1000
and start with the negative reward $r_n$ set to 0, 
decreasing it from the 1000th iteration onward on with a linear slope of -1e-5.
We fix the \textit{softmax} temperature to $100$.
We train for 5k iterations, which takes only 5h on a single Nvidia RTX 2080Ti.

Only after the keypoint \textit{detector} training is complete,
the learned keypoints can be used to train the \textit{descriptor} network. 
Consequently, the \textit{descriptor} network can concentrate solely 
on distinguishing the areas surrounding the detected keypoints and ignore all other regions.
We set the triplet margin $m$ to 0.5, the descriptor dimensionality $d$ to 128 
and train with the same batch size and learning rate as above.
The anchor-positive pairs are built from keypoint pairs that projects closer than 3px. 
We found it beneficial 
to start by sampling random negatives instead of the hardest,
lowering this random sampling probability exponentially and setting it to 0 after 10k iterations.
We train for 90k iterations, which takes around 24h on a single Nvidia RTX 2080Ti.

\section{Experiments}
\label{sec:experiments}
We evaluate our method and compare with other state-of-the-art feature extraction methods via three different experiments.
The first experiment regards only the keypoint \textit{detector} and is meant to evaluate 
its ability to find repeatable keypoints under in-plane rotation.
The other two experiments regard two commonly used benchmarks in the literature 
and test the performances of keypoints and descriptors jointly.
The metrics evaluated are the following:
\begin{itemize}
    \setlength\itemsep{0.0em}
    \item \textbf{Repeatability}: 
        measures the capability of a method to detect the same keypoints in a pair of images
        depicting the same scene. 
        It is computed as the fraction of keypoints from the first image whose projection
        into the second image has at least one keypoint from the second image
        within \textit{T} pixel distance \cite{superpoint}.
    \item \textbf{MMA}:
        The \textit{Mean Matching Accuracy} is computed as the number of correct matches
        (up to \textit{T} pixels) over the total number of matches proposed \cite{d2-net}.
    \item \textbf{MS}:
        The \textit{Matching Score} is computed as the number of correct matches
        (up to \textit{T} pixels) over the average number of detected keypoints in the overlap area \cite{superpoint}.
    \item \textbf{Homography accuracy AUC}:
        It is computed as the Area Under the Curve of the fraction of recovered homographies
        that have a corner error \cite{superpoint} below \textit{T} pixels.
        The corner error is computed as the average distance between the reference image corners
        and the corners of the source image reprojected using the homography estimated from the matches.
    \item \textbf{mAA}:
        The \textit{mean Average Accuracy} is computed as the 
        \textit{Area Under the Curve} of the fraction of recovered relative camera poses
        with an error within a specified threshold \cite{imb}.
\end{itemize}
During inference, 
our ReCNN architecture comes at no additional cost compared to a standard CNN,
and the probabilistic \textit{sequential sampling} is replaced with
a Non-Maximum suppression with a 3px radius and extracts the N keypoints with highest score.
On \textit{Hpatches} and \textit{IMB} we run both our networks on a 5-level image pyramid with
scale factor $1/\sqrt{2}$.

\begingroup
\begin{table}[t]
\begin{adjustbox}{width=1\linewidth}
\setlength{\tabcolsep}{4 pt}
\renewcommand{\arraystretch}{1.2}
\renewcommand\cellalign{cc}
\centering
\begin{tabular}{lc|ccc|ccc|ccc|c}
\cline{2-12}
    \multicolumn{1}{c}{} & 
    \multicolumn{1}{c|}{\multirow{2}{*}{Method}} & 
    \multicolumn{3}{c|}{Rep $\uparrow$} & 
    \multicolumn{3}{c|}{MMA $\uparrow$} & 
    \multicolumn{3}{c|}{MS $\uparrow$} &
    Hom. Acc. \\
    \multicolumn{1}{c}{} & 
    \multicolumn{1}{c|}{} & 
    @1px & @2p & @3px & 
    @1px & @2p & @3px & 
    @1px & @2px & @3px &
    AUC@3px \\
\cline{2-12}
    \multirow{5}{*}{\rotatebox[origin=c]{90}{standard}} 
    & DISK \cite{disk} &
        0.29 & 0.45 & 0.54 & 
        \bf{0.45} & \bf{0.67} & \bf{0.76} & 
        \bf{0.28} & \bf{0.40} & \bf{0.45} & 
        0.438 \\
    & SuperPoint \cite{superpoint} &
        0.23 & 0.41 & 0.54 & 
        0.31 & 0.51 & 0.62 & 
        0.19 & 0.32 & 0.38 & 
        0.413 \\
    & R2D2 \cite{r2d2} &
        \bf{0.31} & \uf{0.50} & \uf{0.60} & 
        0.34 & \uf{0.61} & \uf{0.74} & 
        0.16 & 0.27 & 0.32 & 
        \uf{0.441} \\
    & REKD \cite{REKD} &
        0.22 & 0.42 & 0.54 & 
        0.26 & 0.49 & 0.62 & 
        0.17 & 0.31 & 0.39 & 
        0.416 \\
    & S-TREK (ours) &
        \bf{0.31} & \bf{0.51} & \bf{0.61} & 
        \uf{0.35} & 0.58 & 0.71 & 
        \uf{0.22} & \uf{0.35} & \uf{0.42} & 
        \bf{0.449} \\
\cline{2-12}
    \multirow{5}{*}{\rotatebox[origin=c]{90}{$\pm$ 20°}} 
    & DISK \cite{disk} &
        0.23 & 0.42 & 0.52 & 
        \bf{0.36} & \bf{0.62} & \bf{0.71} & 
        \bf{0.20} & \bf{0.33} & \bf{0.38} & 
        \uf{0.329} \\
    & SuperPoint \cite{superpoint} &
        0.17 & 0.37 & 0.51 & 
        0.22 & 0.45 & 0.57 & 
        0.13 & 0.27 & 0.34 & 
        0.289 \\
    & R2D2 \cite{r2d2} &
        \bf{0.27} & \bf{0.48} & \uf{0.57} & 
        0.27 & 0.55 & 0.68 & 
        0.09 & 0.18 & 0.22 & 
        0.293 \\
    & REKD \cite{REKD} &
        0.17 & 0.38 & 0.51 & 
        0.20 & 0.44 & 0.58 & 
        0.12 & 0.26 & 0.34 & 
        0.224 \\
    & S-TREK (ours) &
        \bf{0.27} & \bf{0.48} & \bf{0.58} & 
        \uf{0.30} & \uf{0.56} & \uf{0.67} & 
        \uf{0.17} & \uf{0.30} & \uf{0.36} & 
        \bf{0.336} \\
\cline{2-12}
    \multirow{5}{*}{\rotatebox[origin=c]{90}{$\pm$ 45°}} 
    & DISK \cite{disk} &
        0.18 & 0.36 & 0.47 & 
        \uf{0.21} & 0.37 & 0.43 & 
        \uf{0.11} & 0.18 & 0.21 & 
        0.182 \\
    & SuperPoint \cite{superpoint} &
        0.14 & 0.33 & 0.47 & 
        0.16 & 0.33 & 0.42 & 
        0.09 & 0.19 & 0.24 & 
        0.181 \\
    & R2D2 \cite{r2d2} &
        \uf{0.22} & \uf{0.42} & \bf{0.53} & 
        0.16 & 0.33 & 0.40 & 
        0.05 & 0.09 & 0.11 & 
        0.165 \\
    & REKD \cite{REKD} &
        0.16 & 0.36 & 0.48 & 
        0.17 & \uf{0.39} & \uf{0.51} & 
        0.10 & \uf{0.22} & \bf{0.29} & 
        \uf{0.184} \\
    & S-TREK (ours) &
        \bf{0.25} & \bf{0.44} & \bf{0.53} & 
        \bf{0.24} & \bf{0.45} & \bf{0.54} & 
        \bf{0.13} & \bf{0.23} & \uf{0.28} & 
        \bf{0.248} \\
\cline{2-12}
\end{tabular}
\vspace{-1.5em}
\end{adjustbox}
\caption{Comparison on HPatches - keypoints budget 2048.}
\label{tab:hpatches}
\vspace*{-2.0mm}
\end{table}
\endgroup

\subsection{Repeatability under rotations}
\label{sec:rotations}
We replicate the experiment conducted in \cite{REKD}. 
Specifically, we select the first ten images from \textit{HPatches} \cite{hpatches},
we generate in-plane rotated versions of these images 
with an increment of 1° from 0° to 360° and
we center-crop them to size 224$\times$224.
As in the original experiment \cite{REKD}, we apply a light gaussian noise to each rotated image.
We set a budget of 50 keypoints and run all the 
learnt methods single-scale.
Keypoints are extracted from each image and the \textit{repeatability} is computed
against the angle 0° image.
The results are depicted in Figure~\ref{fig:rotation_experiment}, where
S-TREK obtains high and very stable repeatability at any angle, showing only minor oscillations.
REKD \cite{REKD}, despite employing a similar rotation-equivariance backbone,
shows a more pronounced oscillation.
While SIFT \cite{sift} has lower but consistent repeatability, 
ORB \cite{orb} displays remarkable repeatability values but also more oscillation.
Finally, while SuperPoint \cite{superpoint} shows modest robustness to rotation,
the other deep methods perform poorly with angles higher than 90°, 
with DISK \cite{disk} reaching values close to zero at 180°.
Additional experiments on this dataset are performed in Sec~\ref{sec:ablation_rotation}.

\begingroup
\newcommand{\cc}{\cellcolor{blue!25}}
\begin{table*}[t]
\begin{adjustbox}{width=1\textwidth}
\setlength{\tabcolsep}{4 pt}  
\renewcommand{\arraystretch}{1.2}  
\centering
\begin{tabular}{cc|cccccccccc|cccccccccc|ccccccccccc|}
\cline{2-32}
    \multicolumn{1}{c}{} & 
    \multicolumn{1}{c|}{\multirow{2}{*}{Method}} & 
    \multicolumn{10}{c|}{Repeatability@3px $\uparrow$} & 
    \multicolumn{10}{c|}{N. matches inliers $\uparrow$} & 
    \multicolumn{10}{c}{mAA@10 $\uparrow$} \\
    \multicolumn{1}{c}{} & 
    & 
    BM & FLC & LM & LB & MC & MR & PSM & SF & SPC & \cc\textbf{AVG} &
    BM & FLC & LM & LB & MC & MR & PSM & SF & SPC & \cc\textbf{AVG} &
    BM & FLC & LM & LB & MC & MR & PSM & SF & SPC & \cc\textbf{AVG} \\
\cline{2-32}
    \multirow{5}{*}{\rotatebox[origin=c]{90}{Standard}}
    & DISK \cite{disk} &
        \bf{0.58} & \uf{0.42} & 0.39 & \bf{0.47} & \uf{0.53} & 0.43 & \uf{0.36} & 0.38 & \uf{0.47} & \cc0.448 &
        \bf{572}  & \bf{400}  & 327  & \bf{350}  & \bf{556}  & \bf{393}  & \bf{265}  & \bf{367}  & \bf{408}  & \cc\bf{404}   &
        \bf{0.41} & \bf{0.69} & 0.59 & \bf{0.58} & \bf{0.53} & \bf{0.38} & \bf{0.26} & \bf{0.58} & \bf{0.59} & \cc\bf{0.512} \\
    & SuperPoint \cite{superpoint} &
        0.42 & 0.36 & 0.37 & 0.35 & 0.38 & 0.41 & 0.30 & 0.35 & 0.34 & \cc0.364 &
        113  & 122  & 115  & 101  & 102  & 130  & 48   & 174  & 82   & \cc110   &
        0.23 & 0.44 & 0.46 & 0.33 & 0.22 & 0.20 & 0.14 & 0.38 & 0.24 & \cc0.295 \\
    & R2D2 \cite{r2d2} &
        0.50 & 0.40 & 0.36 & 0.38 & \uf{0.53} & \bf{0.50} & 0.34 & \uf{0.43} & 0.43 & \cc0.429 & 
        215  & 192  & 220  & 148  & 229  & 271  & 150  & 217  & 172  & \cc202   & 
        0.25 & 0.60 & \uf{0.60} & 0.43 & 0.33 & 0.30 & 0.18 & 0.42 & 0.40 & \cc0.390 \\
    & REKD \cite{REKD} &
        \uf{0.54} & \bf{0.43} & \bf{0.48} & 0.40  & 0.49 & 0.45 & \bf{0.41} & 0.41 & 0.44 & \cc\uf{0.450} &
        284  & 298  & \bf{490}  & 217  & 273  & 310  & 212  & 325  & 254  & \cc296   &
        0.26 & 0.66 & \bf{0.66} & 0.41 & 0.37 & 0.28 & 0.18 & 0.49 & 0.50 & \cc0.423 \\
    & S-TREK (ours) &
        \uf{0.54} & 0.41 & \uf{0.47} & \uf{0.43} & \bf{0.54} & \uf{0.47} & 0.34 & \bf{0.46} & \bf{0.48} & \cc\bf{0.460} &
       \uf{325} & \uf{322} & \uf{419} & \uf{245} & \uf{377} & \uf{319} & \uf{222} & \uf{345} & \uf{323} & \cc\uf{322} &
       \uf{0.33} & \uf{0.67} & 0.53 & \uf{0.47} & \uf{0.43} & \uf{0.34} & \uf{0.23} & \uf{0.54} & \uf{0.58} & \cc\uf{0.458} \\
\cline{2-32}
    \multirow{5}{*}{\rotatebox[origin=c]{90}{$\pm$ 20°}}
    & DISK \cite{disk} &
        \bf{0.55} & 0.39 & 0.39 & \bf{0.47} & \uf{0.53} & 0.43 & \uf{0.36} & 0.37 & \bf{0.45} & \cc0.438 &
        \uf{288}  & 198  & 147  & 195  & \uf{315}  & 224  & 150  & 206  & 221  & \cc216   &
        \uf{0.16} & 0.44 & 0.39 & \uf{0.37} & \uf{0.35} & 0.22 & \uf{0.17} & 0.38 & 0.40  & \cc0.320 \\
    & SuperPoint \cite{superpoint} &
        0.39 & 0.35 & 0.31 & 0.34 & 0.37 & 0.39 & 0.29 & 0.33 & 0.31 & \cc0.342 &
        39   & 42   & 34   & 44   & 42   & 51   & 18   & 68   & 29   & \cc41 &
        0.13 & 0.23 & 0.27 & 0.22 & 0.13 & 0.11 & 0.06 & 0.24 & 0.12 & \cc0.168 \\
    & R2D2 \cite{r2d2} &
        0.45 & 0.40  & 0.36 & 0.39 & 0.52 & \bf{0.49} & 0.34 & \uf{0.43} & 0.42 & \cc0.422 &
        102  & 105  & 110  & 89   & 126  & 140  & 82   & 110  & 89   & \cc106   &
        0.11 & 0.38 & 0.36 & 0.27 & 0.24 & 0.19 & 0.13 & 0.26 & 0.25 & \cc0.243 \\
    & REKD \cite{REKD} &
        0.49 & \bf{0.42} & \bf{0.46} & 0.42 & 0.50  & 0.46 & \bf{0.40}  & 0.41 & 0.43 & \cc\uf{0.443} & 
        249  & \uf{267}  & \bf{386}  & \uf{219}  & 291  & \bf{273}  & \uf{194}  & \bf{291}  & \uf{226}  & \cc\uf{266}   &
        \uf{0.16} & \uf{0.56} & \bf{0.58} & 0.36 & 0.32 & \uf{0.23} & \uf{0.17} & \uf{0.44} & \uf{0.41} & \cc\uf{0.359} \\
    & S-TREK (ours) &
        \uf{0.52} & \uf{0.41} & \uf{0.43} & \uf{0.46} & \bf{0.54} & \uf{0.47} & 0.35 & \bf{0.45} & \bf{0.45} & \cc\bf{0.453} &
        \bf{308} & \bf{288} & \uf{329} & \bf{252} & \bf{348} & \uf{258} & \bf{196} & \uf{280} & \bf{272} & \cc\bf{281} &
        \bf{0.18} & \bf{0.59} & \uf{0.46} & \bf{0.42} & \bf{0.38} & \bf{0.28} & \bf{0.20} & \bf{0.47} & \bf{0.48} & \cc\bf{0.384} \\
\cline{2-32}
    \multirow{5}{*}{\rotatebox[origin=c]{90}{$\pm$ 45°}}
    & DISK \cite{disk} &
        \uf{0.48} & 0.35 & \bf{0.37} & \uf{0.42} & 0.48 & 0.39 & 0.32 & 0.31 & 0.41 & \cc0.392 &
        125  & 92   & 68   & 80   & 138  & 109  & 69   & 96   & 86   & \cc96    &
        0.07 & 0.19 & 0.19 & 0.14 & 0.16 & 0.10  & 0.07 & 0.18 & 0.15 & \cc0.139 \\
    & SuperPoint \cite{superpoint} &
        0.37 & 0.33 & 0.30 & 0.31 & 0.35 & 0.36 & 0.28 & 0.31  & 0.30  & \cc0.323 &
        18   & 18   & 18   & 20  & 18   & 25   & 10   & 27    & 14   & \cc19    &
        0.04 & 0.08 & 0.11 & 0.07 & 0.04 & 0.04 & 0.02 & 0.09 & 0.04 & \cc0.059 \\
    & R2D2 \cite{r2d2} &
        0.39 & 0.38 & 0.32 & 0.36 & 0.49 & \bf{0.47} & 0.32 & 0.39 & 0.40  & \cc0.391 &
        56   & 53   & 53   & 45   & 65   & 71   & 43   & 56   & 43   & \cc54    & 
        0.04 & 0.15 & 0.16 & 0.10 & 0.10   & 0.08 & 0.06 & 0.11 & 0.08 & \cc0.098 \\
    & REKD \cite{REKD} &
        0.46 & \bf{0.41} & \bf{0.40}  & 0.39 & \uf{0.50}  & \uf{0.45} & \bf{0.37} & \uf{0.40}  & \uf{0.42} & \cc\uf{0.422} &
        \uf{178}  & \bf{178}  & \bf{207}  & \uf{146}  & \bf{212}  & \bf{186}  & \bf{138}  & \bf{184}  & \uf{145}  & \cc\bf{175}   &
        \uf{0.09} & \bf{0.35} & \bf{0.36} & \uf{0.20}  & \uf{0.21} & \uf{0.13} & \bf{0.13} & \bf{0.28} & \uf{0.21} & \cc\uf{0.218} \\
    & S-TREK (ours) &
        \bf{0.49} & \bf{0.41} & 0.36 & \bf{0.44} & \bf{0.52} & 0.44 & \uf{0.33} & \bf{0.43} & \bf{0.43} & \cc\bf{0.428} &
        \bf{201} & \uf{176} & \uf{160} & \bf{150} & \uf{205} & \uf{150} & \uf{114} & \uf{147} & \bf{154} & \cc\uf{162} & 
        \bf{0.10} & \bf{0.35} & \uf{0.27} & \bf{0.21} & \bf{0.24} & \bf{0.16} & \bf{0.13} & \bf{0.28} & \bf{0.26} & \cc\bf{0.222} \\
\cline{2-32}
\end{tabular}
\end{adjustbox}
\vspace*{-1.5mm}
\caption{Experiments on Image Matching Benchmark \cite{imb}. The keypoints budget is 2048.}
\label{tab:imb}
\vspace*{-2.5mm}
\end{table*}
\endgroup

\subsection{HPatches}
\label{sec:hpatches}
\textit{HPatches} \cite{hpatches} is a dataset composed by two sets of pictures:
planar scenes captured from different angles
and static photos captured in different lighting conditions.
We run our evaluation using the 108 scenes subset of \cite{d2-net}.
Each scene is composed by one reference image and five source images;
all the metrics are computed between the reference and all the sources.
To test the method stability with respect to rotation,
we generate two additional versions of the whole dataset applying random rotations,
sampled uniformly between $\pm$20° and $\pm$45° to each source image.
Each rotated image is cropped to match
the rectangle with the largest area possible to avoid any border artifacts.
This makes the rotated version of the benchmark even more challenging 
due to the lower overlap between each source and reference image.
We compute the relative homography using the OpenCV \texttt{findHomography} function (10k iterations) 
with multiple RANSAC thresholds (0.125, 0.25, 0.5, 0.75, 1.0, 1.5, 2.0, 2.5, 3.0).
For each method, we compute the homography accuracy AUC at each RANSAC threshold and then pick the best value.
For a more fair evaluation, we run all the multi-scale methods starting from the original image resolution,
fix the keypoints budget to 2048 and always use the same Mutual-Nearest-Neighbor matching strategy.

The results of our evaluation are reported in Table~\ref{tab:hpatches}.
We can observe that the S-TREK detector holds the best repeatability values at every threshold on each rotation set, 
thus validating our proposed detector architecture and training scheme.
Regarding MMA and MS, DISK \cite{disk} is the clear winner for the standard and $\pm$20° sets, 
with S-TREK following it closely on the $\pm$20° and overtaking it on the $\pm$45° set.
For the homography recovery task, MMA is not crucial,
as most of the wrong proposed matches are filtered out by RANSAC.
While MS is a better predictor for the homography accuracy, 
it does not capture how the matches are distributed in the image, 
which is of primal importance for a precise homography recovery.
The high homography accuracy obtained by S-TREK
backs our \textit{``detect, then describe''} approach and
confirms our method capabilities to find repeatable, 
reliable and well distributed local features.

\subsection{Image Matching Benchmark}
We evaluate the performance of local features for the task of \textit{stereo pose recovery} 
on the restricted keypoint category (2048 keypoints per image) 
of the Image Matching Benchmark \cite{imb} phototourism set.
The benchmark consists of 9 scenes of famous tourism sites, each comprising 100 images with various degrees of overlaps.
The online evaluation server has recently been disabled.
For this reason, all the evaluations reported on this paper for methods that were not available 
in the online leaderboard have been run locally.
Similarly to the previous experiment in Sec~\ref{sec:hpatches}, 
we generated two additional instances of the benchmark by applying random $\pm$20° and $\pm$45° rotations,
which also in this case become more challenging due to the decreased overlap between images.
We run all the methods without any additional outlier rejection stage 
and using the Mutual-Nearest-Neighbor matcher.
Where available, we use the matcher parameters and RANSAC threshold suggested by the authors.
Specifically, for S-TREK we use a minimum matching score of 0.5 and a RANSAC threshold of 1.0 px.

The results for \textit{repeatability}, \textit{number of inlier matches} and \textit{mAA@10} for each scene in the dataset are reported in Table~\ref{tab:imb}.
Similarly to the \textit{HPatches} experiments, 
S-TREK is the method with the highest repeatability in all the benchmark sets.
While DISK \cite{disk} obtains the best \textit{number of inlier matches} and \textit{mAA@10} in the standard instance,
S-TREK follows it closely with a competitive 0.458 \textit{mAA@10}.
On the rotated versions of the benchmark DISK falls behind S-TREK and REKD,
which achieve similar performance regarding the \textit{number of inlier matches}.
S-TREK provides the best \textit{mAA@10} for both angle sets,
confirming again the effectiveness of our approach.
Figure~\ref{fig:imb} shows a qualitative comparison between the best performing methods
where S-TREK is the only method which provides well distributed matches in both standard, 
and $\pm$45° benchmark instances.
\vspace*{-1.0mm}

\section{Ablation Studies}

\subsection{Rotation equivariant architectures comparison}
\label{sec:ablation_rotation}

\begin{figure}[t]
\begin{center}
\includegraphics[width=\linewidth]{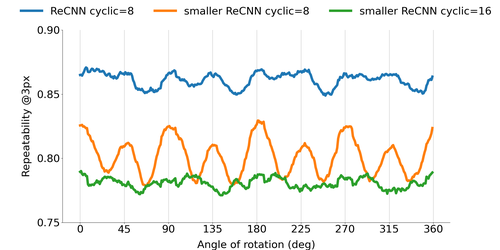}
\end{center}
\vspace*{-4mm}
\caption{
Comparison between keypoint detector architectures in terms of keypoint repeatability as a function of image rotation angle. ReCNN $\text{cyclic}=8$ is the S-TREK detector.
}
\vspace*{-2mm}
\label{fig:ablation_rotation}
\vspace*{-1mm}
\end{figure}

We repeat the experiment from Section~\ref{sec:rotations} 
using three different keypoint detector architectures:
\begin{itemize}
    \setlength\itemsep{0.0em}
    \item \textbf{ReCNN cyclic 8}: Baseline S-TREK \textit{detector} architecture 
    ($\sim$20k learnable params).
    \item \textbf{smaller ReCNN cyclic 8}: shallow ReCNN with cyclicity 8
    ($\sim$5k learnable params).
    \item \textbf{smaller ReCNN cyclic 16}: shallow ReCNN with cyclicity 16
    ($\sim$10k learnable params).
\end{itemize}
The results are shown in Figure~\ref{fig:ablation_rotation}.
It can be observed that 
the smaller ReCNN with cyclicity 8 exhibits 
a repeatability with lower mean and larger fluctuations than the baseline.
Increasing the cyclicity to 16 smooths out the fluctuations but does not increase the mean repeatability.
The baseline, i.e. S-TREK detector, is able to reduce the fluctuations without requiring higher cyclicity
thanks to the deeper expressivity given by 
the larger number of 
convolutional layers
(7 vs 4).

\subsection{Different losses and architectures}
We compare our sequential and reinforcement learning inspired keypoint training 
against the commonly used \textit{peaky} and \textit{similarity} losses \cite{r2d2, mdnet},
which need to be used in conjunction and are 
applied to the detection heatmap.
The \textit{peaky} loss encourages local peaks and is controlled by a window-size parameter, 
while the \textit{similarity} loss promotes similar detection heatmaps in a dense manner.

To analyse the convergence properties of the different training formulations,
we train on a synthetic dataset generated by drawing random grayscale lines
plus random gaussian noise.
Every pair of images is obtained applying two random homography warpings
to a generated lines image.

\begin{figure}
     \newcommand{\mysize}{0.30\linewidth}
     \begin{subfigure}[b]{\mysize}
         \includegraphics[width=\linewidth]{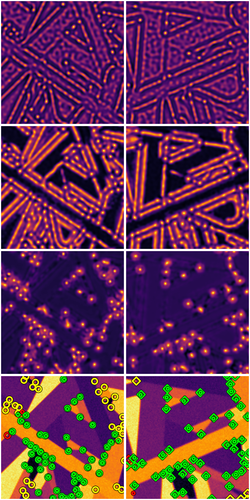}
         \caption{N. samples 200}
     \end{subfigure}
     \begin{subfigure}[b]{\mysize}
         \includegraphics[width=\linewidth]{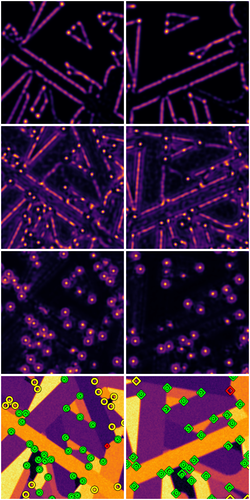}
         \caption{N. samples 50}
     \end{subfigure}
     \begin{subfigure}[b]{\mysize}
         \includegraphics[width=\linewidth]{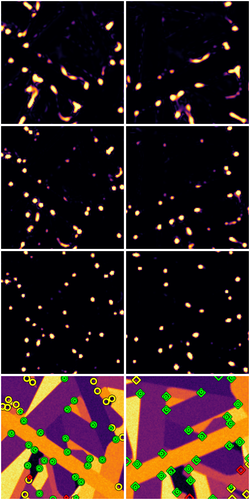}
         \caption{Peaky W=64}
     \end{subfigure}
     \begin{subfigure}[b]{0.01\textwidth}
        \begin{tikzpicture}
          \tikz \fill [white] (0.0,0.0) rectangle (0.1,0.2); 
          \draw[->] (0,5.4) -- (0,2.1);
          \node[rotate=90, font=\footnotesize] at (0.25,3.85) {Training iterations}; 
        \end{tikzpicture}
     \end{subfigure}
     \vspace*{-2.0mm}
     \caption{
     Visual comparison between different training methods on 
     our lines dataset.
     Last row: input images and detected keypoints
     (Green - repeatable, Red - non repeatable, Yellow - non overlapping).
     First three rows: detector heatmaps evolution during training. 
     Our training framework excels at finding
     repeatable points without any direct supervision.
     }
     \label{fig:synthetic}
\end{figure}
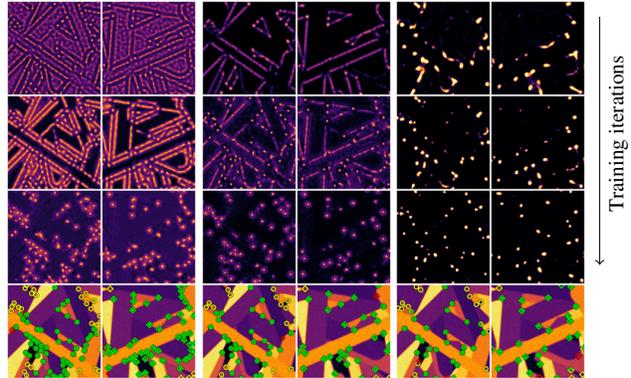

The heatmap evolution during training is shown in Figure~\ref{fig:synthetic} 
and provides insights on the convergence process.
Regardless of the training strategy,
the network starts by highlighting both edges and corners,
progressively favouring corners more.
However, in this process,
the peakyness loss tends to loose good points.
In contrast, 
our training framework remains more stable in terms of number of the detected points, 
regardless of the number of chosen samples.
This is confirmed by the numerical values in Table~\ref{tab:synthetic_dataset}.
The middle section reports
the results of our S-TREK keypoint \textit{detector} trainings using different \textit{peaky} window sizes.
It can be noticed that the number of detected keypoints during inference varies significantly depending on the chosen 
\textit{peaky} window size, ranging from 27 to 60.
The first section 
instead shows the results with a varying number of sequential samples,
for which the number of
keypoints is more stable,
confirming the ability of our training scheme to adapt the number of 
keypoints to the number of existing stable points.

In the last section of the same table we compare the results of training
different standard Convolutional Networks.
The \textit{CNN same channels} refers to a standard CNN with the same number of layers 
and channels shown in Figure~\ref{fig:architecture-det},
while \textit{CNN equivalent} refers to a standard CNN with the same number of layers 
shown in Figure~\ref{fig:architecture-det} where the channel count has been multiplied by the S-TREK cyclicity.
The ReCNN, in conjunction with our training framework,
obtains the highest \textit{max repeatability} values while still keeping a high keypoints count,
confirming the ability of our method to learn to detect the stable points available in the dataset.

\begingroup
\begin{table}[!t]
\begin{adjustbox}{width=\linewidth}
\setlength{\tabcolsep}{3 pt}
\renewcommand{\arraystretch}{1.2}
\renewcommand\cellalign{cc}
\centering
\begin{tabular}{ccccccc}
\hline
    \multicolumn{1}{c}{\multirow{2}{*}{Architecture}} & 
    \multicolumn{1}{c}{\multirow{2}{*}{Training}} & 
    \multicolumn{1}{c}{\multirow{2}{*}{n. kpts}} & 
    \multicolumn{3}{c}{Rep. max. $\uparrow$} &
    \multicolumn{1}{c}{\multirow{2}{*}{\makecell{Learnable \\ Parameters}}} \\
& & & @1px & @2px & @3px & \\
\hline
    \multirow{6}{*}{ReCNN}
      & Sequential n. samples 50 &
        39 & \uf{0.77} & \uf{0.92} & 0.93 &
        ~20k \\
    & Sequential n. samples 100 &
        53 & \bf{0.78} & \bf{0.93} & \bf{0.95} & 
        ~20k \\
    & Sequential n. samples 200 &
        56 & 0.72 & 0.91 & \uf{0.94} & 
        ~20k \\
\cline{2-7}
    & Peaky + Similarity W 32 &
        60 & 0.51 & 0.72 & 0.78 & 
        ~20k \\
    & Peaky + Similarity W 64 &
        36 & 0.58 & 0.82 & 0.88 & 
        ~20k \\
    & Peaky + Similarity W 96 &
        27 & 0.61 & 0.85 & 0.90 &
        ~20k \\
\hline
    CNN same channels & Sequential n. samples 100 &
        19 & 0.61 & 0.85 & 0.90 & 
        ~6k \\
    CNN equivalent & Sequential n. samples 100 &
        38 & 0.68 & 0.90 & 0.92 & 
        ~377k \\
\hline
\end{tabular}
\end{adjustbox}
\vspace*{-1.5mm}
\caption{
Comparison between different training approaches on the validation set
of our 
lines dataset (see Figure~\ref{fig:synthetic}).
}
\label{tab:synthetic_dataset}
\vspace*{-2.0mm}
\end{table}
\endgroup

\section{Conclusion and future works}
This paper presents S-TREK, a local feature extractor method that combines 
a deep keypoint detector that is both translation and rotation equivariant 
with a lightweight deep descriptor extractor.
The proposed training framework and reward formulation, 
inspired by reinforcement learning, 
maximizes the keypoint repeatability score directly.
Moreover, we model the probabilistic keypoint sampling process adopting a 
sequential scheme that overcomes the limitation of previous approaches.
Extensive experiments show that the S-TREK detector often
outperforms the state-of-the-art in the repeatability metric.
Paired with our learnt descriptors, 
S-TREK achieves competitive matching performance, 
especially in applications with strong in-plane rotations. 

A possible direction for future improvements of the S-TREK features could be to incorporate scale equivariance in the network architecture. 
This could further reduce the number of learnable parameters and, therefore,
the amount of data and the computational efforts required by the training.

\textbf{Acknowledgement}: This work has been supported by the FFG, Contract No. 881844: ”Pro2Future”.

\section{Supplementary material}

\begin{figure*}[b!]
     \centering
     \newcommand{\mysize}{0.136\linewidth}
     \centering
     \begin{subfigure}{\mysize}
         \includegraphics[width=0.9\linewidth, trim={0 0 0 1.5cm},clip]{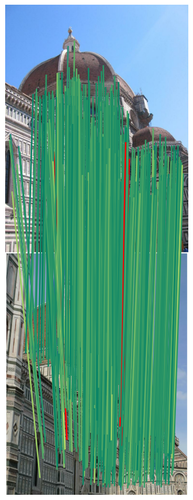}
     \end{subfigure}
     \begin{subfigure}{\mysize}
         \includegraphics[width=0.9\linewidth, trim={0 0 0 1.5cm},clip]{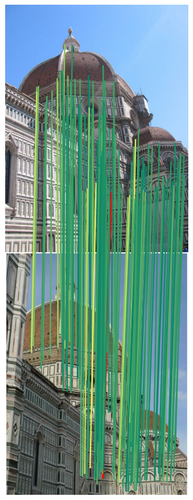}
     \end{subfigure}
     \begin{subfigure}{\mysize}
         \includegraphics[width=0.9\linewidth, trim={0 0 0 1.5cm},clip]{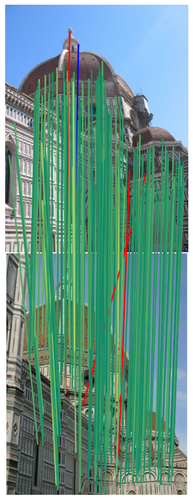}
     \end{subfigure}
     \begin{subfigure}{\mysize}
         \includegraphics[width=0.9\linewidth, trim={0 0 0 1.5cm},clip]{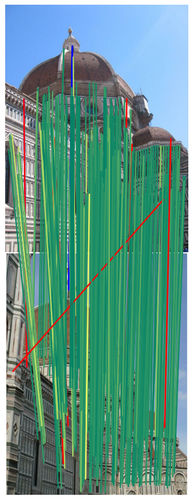}
     \end{subfigure}
     \begin{subfigure}{\mysize}
         \includegraphics[width=0.9\linewidth, trim={0 0 0 1.5cm},clip]{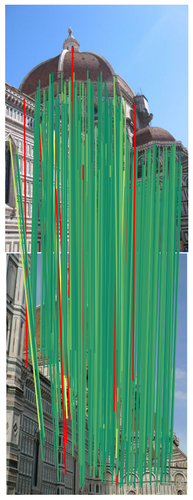}
     \end{subfigure}
     \\
     \begin{subfigure}{\mysize}
         \includegraphics[width=0.9\linewidth, trim={0 0 0 1.5cm},clip]{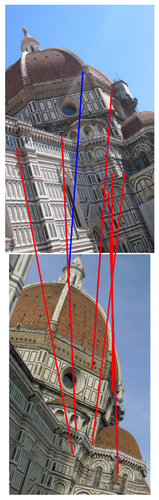}
     \end{subfigure}
     \begin{subfigure}{\mysize}
         \includegraphics[width=0.9\linewidth, trim={0 0 0 1.5cm},clip]{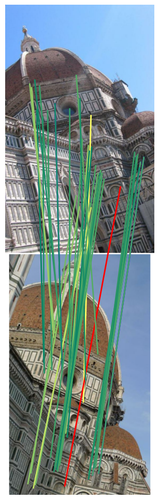}
     \end{subfigure}
     \begin{subfigure}{\mysize}
         \includegraphics[width=0.9\linewidth, trim={0 0 0 1.5cm},clip]{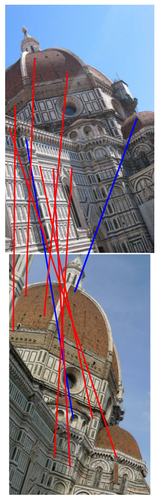}
     \end{subfigure}
     \begin{subfigure}{\mysize}
         \includegraphics[width=0.9\linewidth, trim={0 0 0 1.5cm},clip]{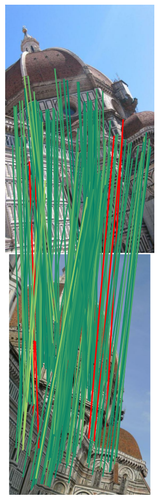}
     \end{subfigure}
     \begin{subfigure}{\mysize}
         \includegraphics[width=0.9\linewidth, trim={0 0 0 1.5cm},clip]{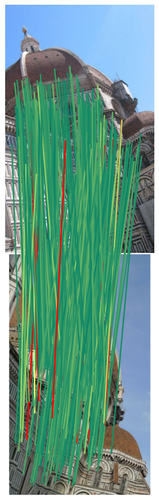}
     \end{subfigure}
     \\
     \begin{subfigure}{\mysize}
         \includegraphics[width=0.9\linewidth, trim={0 0 0 2.5cm},clip]{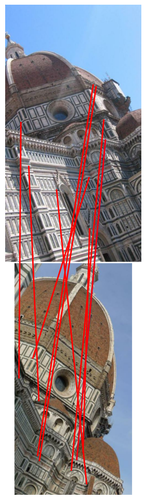}
         \caption{DISK \cite{disk}}
     \end{subfigure}
     \begin{subfigure}{\mysize}
         \includegraphics[width=0.9\linewidth, trim={0 0 0 2.5cm},clip]{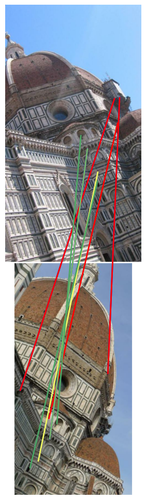}
         \caption{SuperPoint \cite{superpoint}}
     \end{subfigure}
     \begin{subfigure}{\mysize}
         \includegraphics[width=0.9\linewidth, trim={0 0 0 2.5cm},clip]{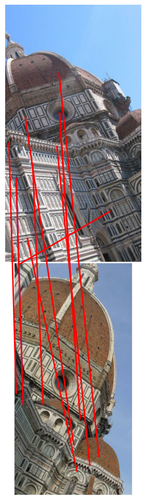}
         \caption{R2D2 \cite{r2d2}}
     \end{subfigure}
     \begin{subfigure}{\mysize}
         \includegraphics[width=0.9\linewidth, trim={0 0 0 2.5cm},clip]{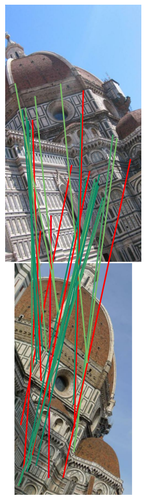}
         \caption{REKD \cite{REKD}}
     \end{subfigure}
     \begin{subfigure}{\mysize}
         \includegraphics[width=0.9\linewidth, trim={0 0 0 2.5cm},clip]{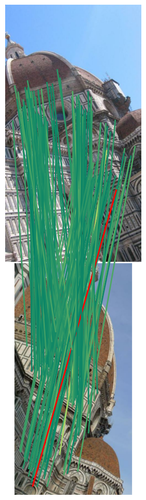}
         \caption{S-TREK (ours)}
     \end{subfigure}
     \vspace{-1mm}
     \caption{
     Qualitative comparison with state-of-the-art feature extraction methods on the Image Matching Benchmark \cite{imb} (top) and on our $\pm$20° and $\pm$45° rotated version of it. 
     RANSAC inlier matches are color coded from green to yellow, representing reprojection errors equal to zero and 5px, respectively; the outlier matches are in red.
     }
     \label{fig:imb_extra0}
\end{figure*}

\begin{figure*}[h]
     \centering
     \newcommand{\mysize}{0.187\linewidth}
     \centering
     \begin{subfigure}{\mysize}
         \includegraphics[width=\linewidth]{media/experiments/imb/00040_DISK.png}
     \end{subfigure}
     \begin{subfigure}{\mysize}
         \includegraphics[width=\linewidth]{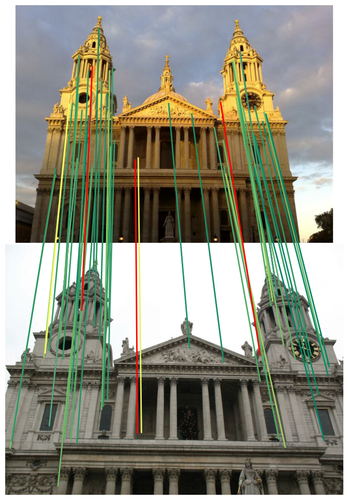}
     \end{subfigure}
     \begin{subfigure}{\mysize}
         \includegraphics[width=\linewidth]{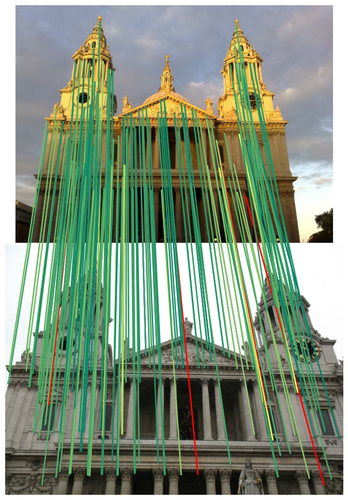}
     \end{subfigure}
     \begin{subfigure}{\mysize}
         \includegraphics[width=\linewidth]{media/experiments/imb/00040_REKD.png}
     \end{subfigure}
     \begin{subfigure}{\mysize}
         \includegraphics[width=\linewidth]{media/experiments/imb/00040_STREK.png}
     \end{subfigure}
     \\
     \begin{subfigure}{\mysize}
         \includegraphics[width=\linewidth]{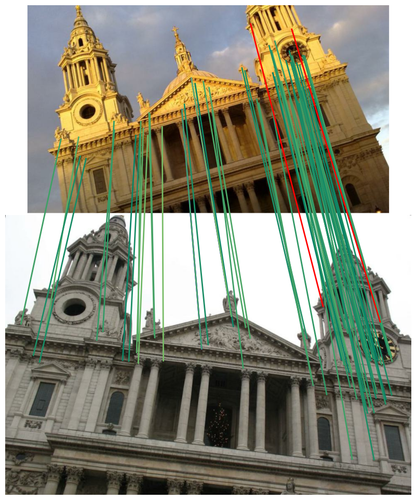}
     \end{subfigure}
     \begin{subfigure}{\mysize}
         \includegraphics[width=\linewidth]{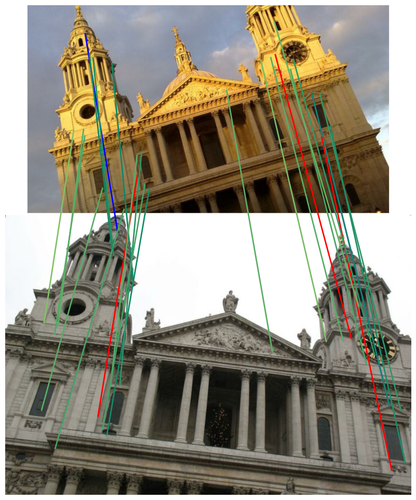}
     \end{subfigure}
     \begin{subfigure}{\mysize}
         \includegraphics[width=\linewidth]{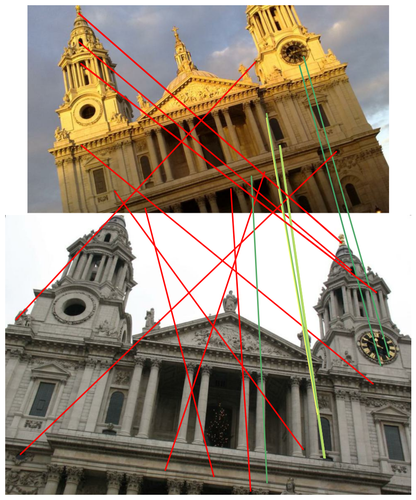}
     \end{subfigure}
     \begin{subfigure}{\mysize}
         \includegraphics[width=\linewidth]{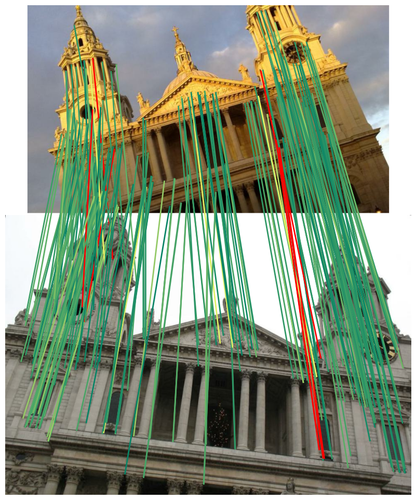}
     \end{subfigure}
     \begin{subfigure}{\mysize}
         \includegraphics[width=\linewidth]{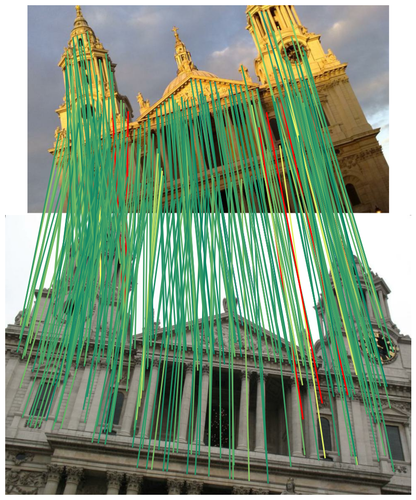}
     \end{subfigure}
     \\
     \begin{subfigure}{\mysize}
         \includegraphics[width=\linewidth]{media/experiments/imb/00040_45_DISK.png}
         \caption{DISK \cite{disk}}
     \end{subfigure}
     \begin{subfigure}{\mysize}
         \includegraphics[width=\linewidth]{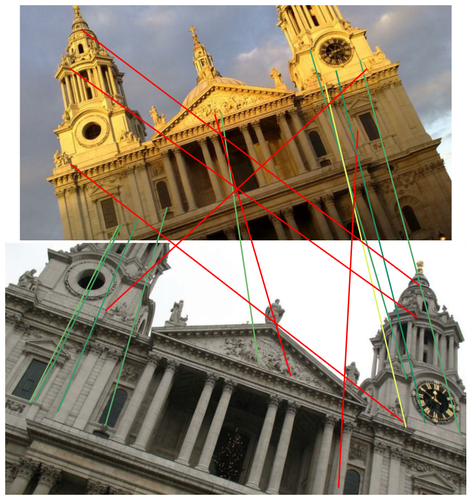}
         \caption{SuperPoint \cite{superpoint}}
     \end{subfigure}
     \begin{subfigure}{\mysize}
         \includegraphics[width=\linewidth]{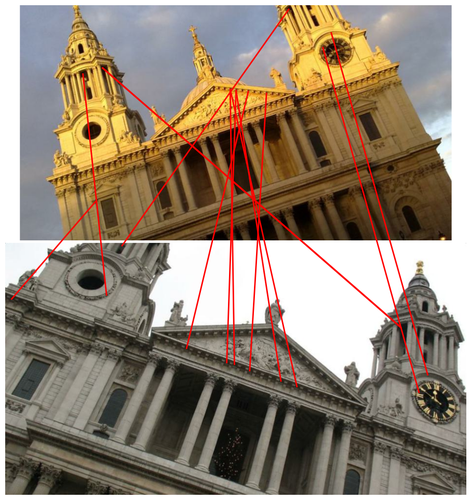}
         \caption{R2D2 \cite{r2d2}}
     \end{subfigure}
     \begin{subfigure}{\mysize}
         \includegraphics[width=\linewidth]{media/experiments/imb/00040_45_REKD.png}
         \caption{REKD \cite{REKD}}
     \end{subfigure}
     \begin{subfigure}{\mysize}
         \includegraphics[width=\linewidth]{media/experiments/imb/00040_45_STREK.png}
         \caption{S-TREK (ours)}
     \end{subfigure}
     \caption{
     Qualitative comparison with state-of-the-art feature extraction methods on the Image Matching Benchmark \cite{imb} (top) and on our $\pm$20° and $\pm$45° rotated version of it. 
     RANSAC inlier matches are color coded from green to yellow, representing reprojection errors equal to zero and 5px, respectively; the outlier matches are in red.
     }
     \label{fig:imb_extra1}
\end{figure*}

\begin{figure*}[h]
     \centering
     \newcommand{\mysize}{0.18\linewidth}
     \centering
     \begin{subfigure}{\mysize}
         \includegraphics[width=\linewidth, trim={0 0cm 0 0.5cm},clip]{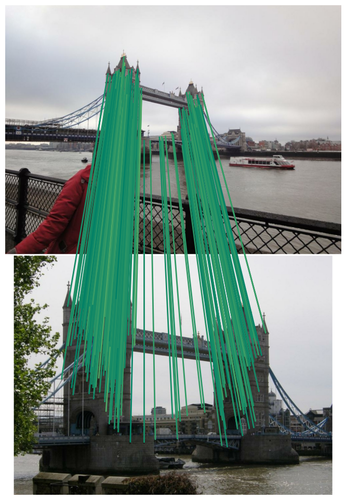}
     \end{subfigure}
     \begin{subfigure}{\mysize}
         \includegraphics[width=\linewidth, trim={0 0cm 0 0.5cm},clip]{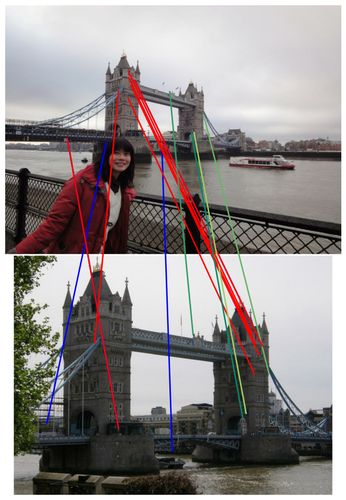}
     \end{subfigure}
     \begin{subfigure}{\mysize}
         \includegraphics[width=\linewidth, trim={0 0cm 0 0.5cm},clip]{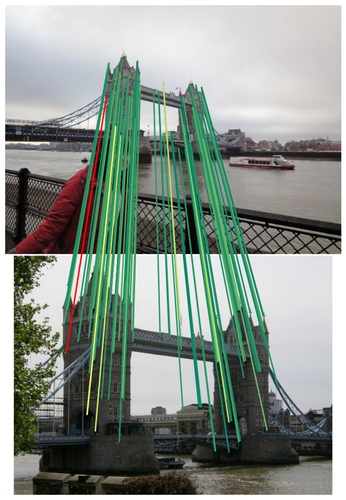}
     \end{subfigure}
     \begin{subfigure}{\mysize}
         \includegraphics[width=\linewidth, trim={0 0cm 0 0.5cm},clip]{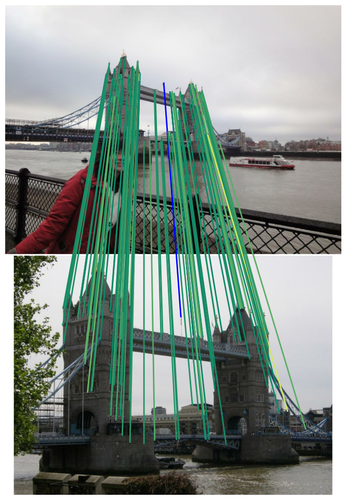}
     \end{subfigure}
     \begin{subfigure}{\mysize}
         \includegraphics[width=\linewidth, trim={0 0cm 0 0.5cm},clip]{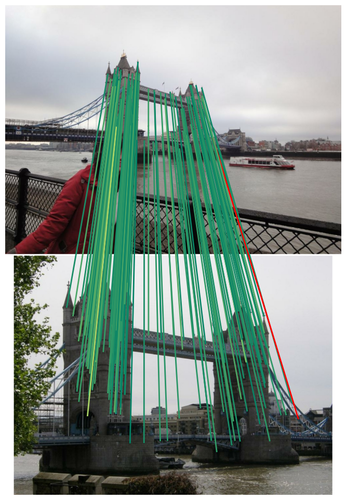}
     \end{subfigure}
     \\
     \begin{subfigure}{\mysize}
         \includegraphics[width=\linewidth, trim={0 0cm 0 0.5cm},clip]{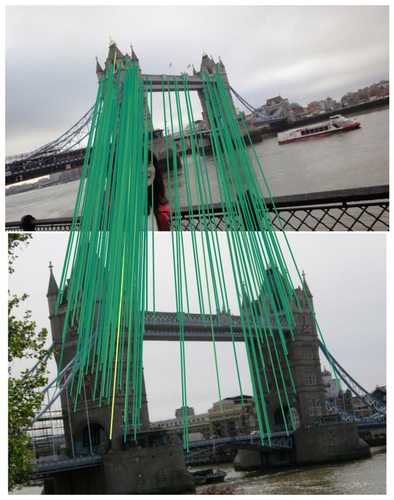}
     \end{subfigure}
     \begin{subfigure}{\mysize}
         \includegraphics[width=\linewidth, trim={0 0cm 0 0.5cm},clip]{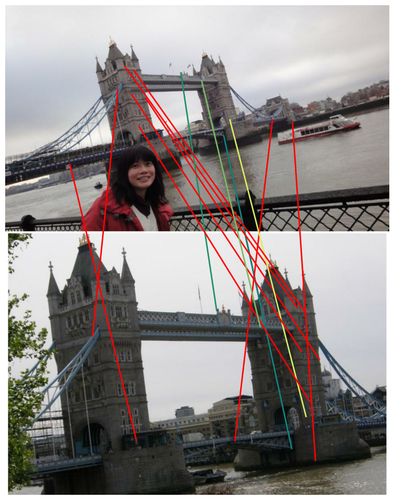}
     \end{subfigure}
     \begin{subfigure}{\mysize}
         \includegraphics[width=\linewidth, trim={0 0cm 0 0.5cm},clip]{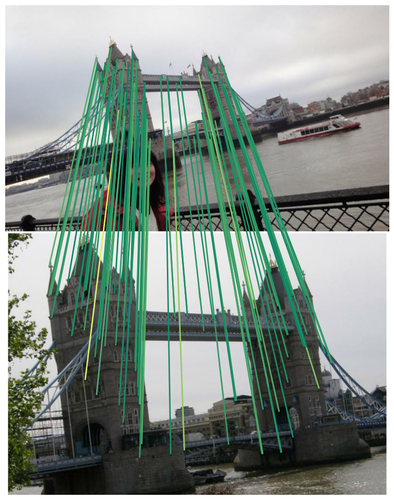}
     \end{subfigure}
     \begin{subfigure}{\mysize}
         \includegraphics[width=\linewidth, trim={0 0cm 0 0.5cm},clip]{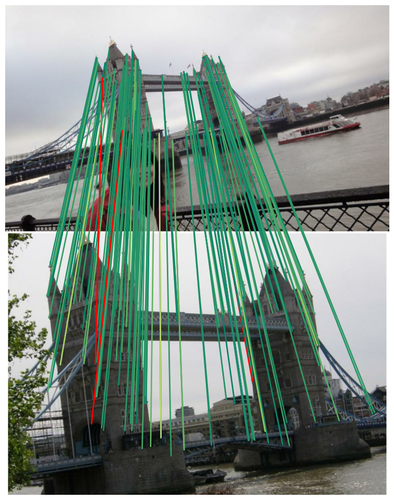}
     \end{subfigure}
     \begin{subfigure}{\mysize}
         \includegraphics[width=\linewidth, trim={0 0cm 0 0.5cm},clip]{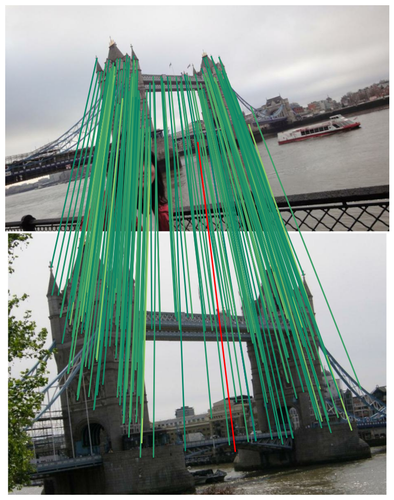}
     \end{subfigure}
     \\
     \begin{subfigure}{\mysize}
         \includegraphics[width=\linewidth, trim={0 0 0 1.5cm},clip]{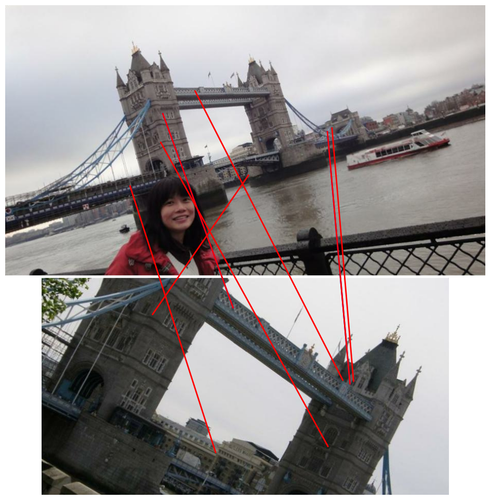}
         \caption{DISK \cite{disk}}
     \end{subfigure}
     \begin{subfigure}{\mysize}
         \includegraphics[width=\linewidth, trim={0 0 0 1.5cm},clip]{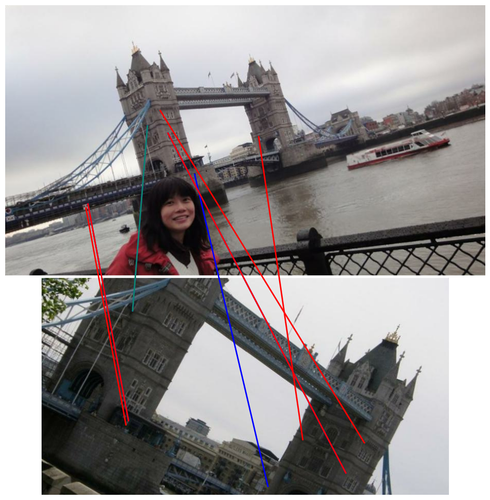}
         \caption{SuperPoint \cite{superpoint}}
     \end{subfigure}
     \begin{subfigure}{\mysize}
         \includegraphics[width=\linewidth, trim={0 0 0 1.5cm},clip]{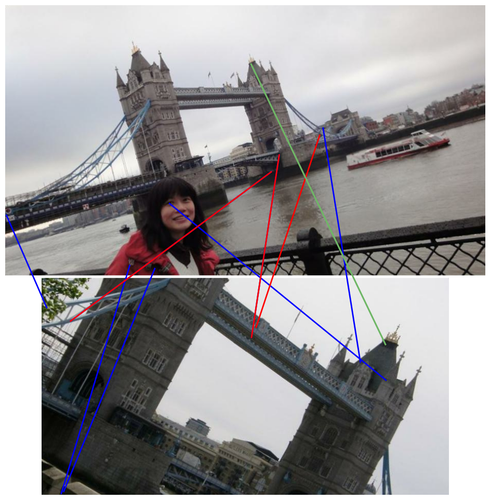}
         \caption{R2D2 \cite{r2d2}}
     \end{subfigure}
     \begin{subfigure}{\mysize}
         \includegraphics[width=\linewidth, trim={0 0 0 1.5cm},clip]{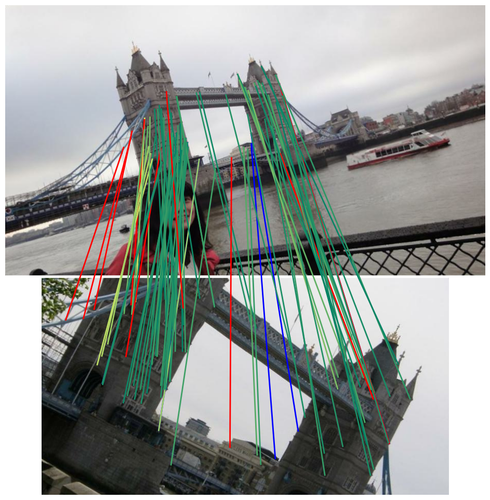}
         \caption{REKD \cite{REKD}}
     \end{subfigure}
     \begin{subfigure}{\mysize}
         \includegraphics[width=\linewidth, trim={0 0 0 1.5cm},clip]{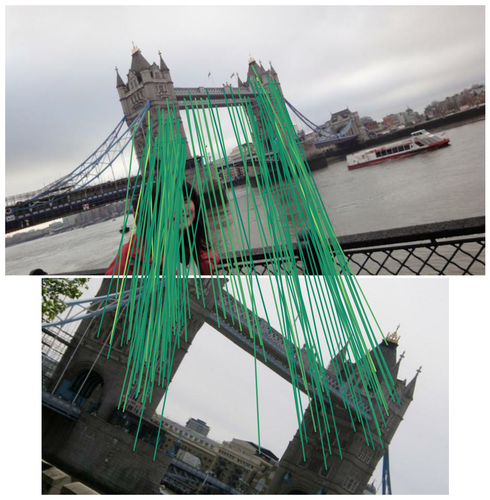}
         \caption{S-TREK (ours)}
     \end{subfigure}
     \caption{
     Qualitative comparison with state-of-the-art feature extraction methods on the Image Matching Benchmark \cite{imb} (top) and on our $\pm$20° and $\pm$45° rotated version of it. 
     RANSAC inlier matches are color coded from green to yellow, representing reprojection errors equal to zero and 5px, respectively; the outlier matches are in red.
     }
     \label{fig:imb_extra2}
\end{figure*}

We include additional qualitative comparisons on \textit{Image Matching Benchmark} \cite{imb}
and our rotated versions with all the tested methods in Figure~\ref{fig:imb_extra0}, Figure~\ref{fig:imb_extra1} and Figure~\ref{fig:imb_extra2}.

\begin{figure*}[t]
     \newcommand{\mysize}{0.16\linewidth}
     \begin{subfigure}[b]{\mysize}
         \includegraphics[width=\linewidth]{media/ablations_synthetic/sample200.png}
         \caption{N. samples 200}
     \end{subfigure}
     \begin{subfigure}[b]{\mysize}
         \includegraphics[width=\linewidth]{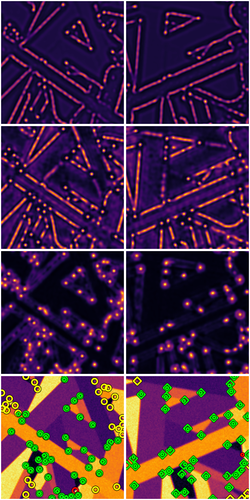}
         \caption{N. samples 100}
     \end{subfigure}
     \begin{subfigure}[b]{\mysize}
         \includegraphics[width=\linewidth]{media/ablations_synthetic/sample50.png}
         \caption{N. samples 50}
     \end{subfigure}
     \begin{subfigure}[b]{\mysize}
         \includegraphics[width=\linewidth]{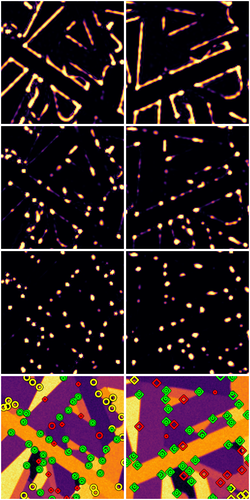}
         \caption{Peaky W=32}
     \end{subfigure}
     \begin{subfigure}[b]{\mysize}
         \includegraphics[width=\linewidth]{media/ablations_synthetic/peaky_n64.png}
         \caption{Peaky W=64}
     \end{subfigure}
     \begin{subfigure}[b]{\mysize}
         \includegraphics[width=\linewidth]{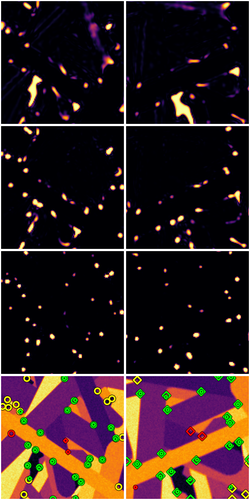}
         \caption{Peaky W=96}
     \end{subfigure}
     \begin{subfigure}[b]{0.01\textwidth}
        \begin{tikzpicture}
          \tikz \fill [white] (0.0,0.0) rectangle (0.1,0.2); 
          \draw[->] (0,5.4) -- (0,2.1);
          \node[rotate=90, font=\footnotesize] at (0.25,3.85) {Training iterations}; 
        \end{tikzpicture}
     \end{subfigure}
     \caption{
     Visual comparison between different training methods on the validation set of the synthetic lines dataset.
     Last row: input images and detected keypoints
     (Green - repeatable, Red - non repeatable, Yellow - non overlapping).
     First three rows: detector heatmaps evolution during training. 
     Our training framework excels at finding
     the repeatable points without any direct supervision.
     }
     \label{fig:synthetic_extra}
\end{figure*}

In Figure~\ref{fig:synthetic_extra} we include an extended version of the main paper figure which shows the detector heatmaps evolution during training.
We include the additional 32 and 96 \textit{Peaky} loss window sizes 
and our training framework run using 100 serial samples.

\FloatBarrier

\clearpage

{\small
\bibliographystyle{ieee_fullname}
\bibliography{arxiv}
}

\end{document}